\documentclass{ieeeaccess}
\usepackage{cite}
\usepackage{amsmath,amssymb,amsfonts}
\usepackage{algorithmic}
\usepackage{graphicx}
\usepackage{textcomp}

\usepackage{booktabs}
\usepackage{multirow}
\usepackage{siunitx}
\usepackage{makecell}
\usepackage{placeins}

\newcommand{\rev}[1]{#1}
\newcommand{\newrev}[1]{#1}

\usepackage{hyperref}
\usepackage{bm}
\makeatletter
\AtBeginDocument{\DeclareMathVersion{bold}
\SetSymbolFont{operators}{bold}{T1}{times}{b}{n}
\SetSymbolFont{NewLetters}{bold}{T1}{times}{b}{it}
\SetMathAlphabet{\mathrm}{bold}{T1}{times}{b}{n}
\SetMathAlphabet{\mathit}{bold}{T1}{times}{b}{it}
\SetMathAlphabet{\mathbf}{bold}{T1}{times}{b}{n}
\SetMathAlphabet{\mathtt}{bold}{OT1}{pcr}{b}{n}
\SetSymbolFont{symbols}{bold}{OMS}{cmsy}{b}{n}
\renewcommand\boldmath{\@nomath\boldmath\mathversion{bold}}}
\makeatother

\def\BibTeX{{\rm B\kern-.05em{\sc i\kern-.025em b}\kern-.08em
    T\kern-.1667em\lower.7ex\hbox{E}\kern-.125emX}}

\usepackage{algorithm}

\begin{document}

\title{\newrev{Towards Test-Time Adaptation for Emergent Language under Complexity Shifts}}
\author{\uppercase{Mohammad Mahdi Samiei}\authorrefmark{1},
\uppercase{Mehdi Jamalkhah}\authorrefmark{2}, and MAHDIEH SOLEYMANI BAGHSHAH\authorrefmark{3}}

\address[1]{Department of Computer Engineering, Sharif University of Technology, Tehran, Iran (e-mail: mm.samiei@sharif.edu)}
\address[2]{Department of Computer Engineering, University of Tehran, Tehran, Iran (e-mail: mehdi.jamalkhah@ut.ac.ir)}
\address[3]{Department of Computer Engineering, Sharif University of Technology, Tehran, Iran (e-mail: soleymani@sharif.edu)}

\markboth
{Mahdi Samiei \headeretal: \newrev{Test-Time Adaptation for Emergent Language under Complexity Shifts}}
{Mahdi Samiei \headeretal: \newrev{Test-Time Adaptation for Emergent Language under Complexity Shifts}}

\corresp{Corresponding author: Mahdieh Soleymani (e-mail: soleymani@sharif.edu).}
\tfootnote{This work has been submitted to the IEEE for possible publication. Copyright may be transferred without notice, after which this version may no longer be accessible.}

\begin{abstract}
Emergent communication research has achieved high coordination in referential games, inducing discrete protocols that resemble language-like symbol systems.
Despite extensive work across diverse emergent-language settings, a key gap remains: what happens when test instances demand finer distinctions or far more confusable distractors than in training?
In emergent-language setups, however,
the sender policy is fixed after training,
leaving it unable to
produce a more informative message when test instances become more complex, despite the expectation we place on the flexibility and adaptability of language.
To address this deployment-time mismatch, we introduce a sender-side test-time adaptation framework
that optimizes message generation using only the sender’s internal self-play objective.
\rev{While this concept represents a general framework, we specifically instantiate and evaluate it using a Vector-Quantized Emergent Language (VQ-EL) architecture,}
\rev{enabling variable-length descriptions} \rev{using inference-time optimization without extra interaction rounds}.
Our experiments demonstrate that, across more complex inputs and more challenging candidate sets, our method
consistently improves communication accuracy and, when needed, benefits from more descriptive messages.
Our findings position test-time adaptation as a practical mechanism for making emergent languages robust to increased deployment-time complexity, offering a novel approach to adapting agents for real-world referential settings where the required message specificity is unpredictable. \rev{Code is available at: \url{https://github.com/MehdiJmlkh/Test-Time-Emergent-Language}.}
\end{abstract}


\titlepgskip=-21pt

\maketitle

\section{Introduction}

Emergent Language (EL) studies how communication protocols can arise between artificial agents that are trained to solve a task in which coordination depends on exchanging messages \cite{lazaridou2020emergent}.
A widely used benchmark for EL is the referential game \cite{Llewelyn_1970}.
In this game, a sender observes a target object (e.g., an image) and produces a discrete message, while a receiver observes a set of candidate objects containing the target and several distractors and must identify the correct target based on the sender’s message \cite{lazaridou2016multi, Havrylov2017EmergenceOL}.
Since the communication channel is discrete, learning meaningful messages is non-trivial and has motivated a large body of work on how languages can emerge under different architectural, optimization, and environmental constraints \cite{Kottur2017NaturalLD, Lazaridou2018EmergenceOL, Lowe2019OnTP}.

\begin{figure*}[t]
  \includegraphics[width=\linewidth]{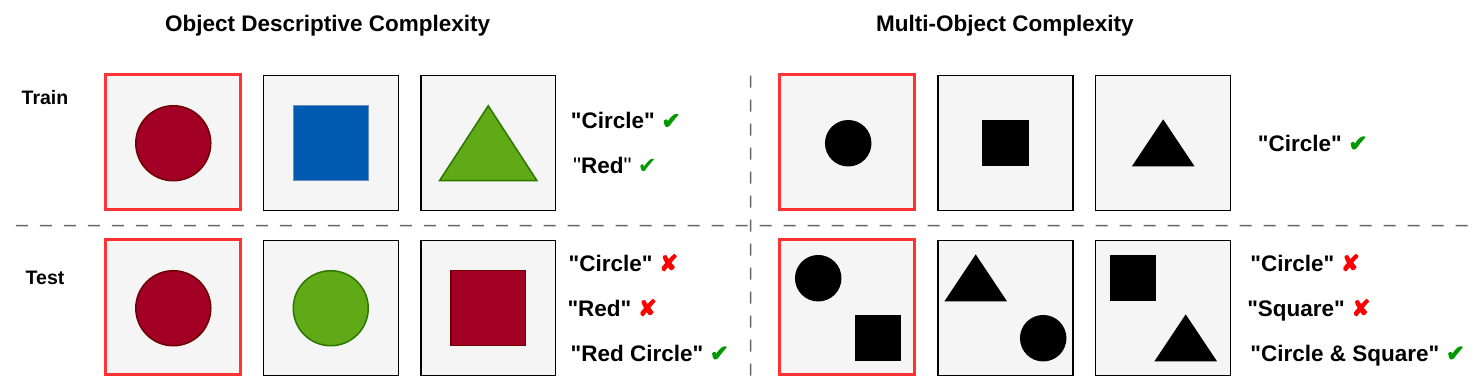}
    \caption{Illustration of increasing object descriptive complexity and multi-object complexity at test time. As complexity grows, a longer message may be required to accurately identify the target (indicated by a red border). The left example is inspired by \cite{zhang2025combinatorial}.}
  \label{fig:complexities}
\end{figure*}

Over the last few years, many studies have proposed training settings designed to regularize or shape emergent communication.
Efforts to achieve emergent languages that evolve in ways analogous to natural languages or exhibit similar linguistic properties include training populations of agents \cite{Gupta2021DynamicPM}, iterated learning or generational dynamics \cite{Ren2020CompositionalLE}, and evaluating or encouraging properties such as compositionality and systematic generalization \cite{chaabouni-etal-2020-compositionality}.
These directions have significantly improved our understanding of how training conditions affect the structure and robustness of emergent languages.
However, an important expectation of language, tied to its compositional nature, has received little attention: what happens when, at deployment, agents encounter test instances that require longer descriptions to distinguish them?
\newrev{The multi-object settings in our study can be understood through the lens of Complexity Out-of-Distribution (Complexity-OoD) generalization \cite{samiei2026bridging}: performance must be maintained when the minimal required solution complexity at test time exceeds that of all training examples. For ImageNet and COCO, we use the more specific description of object-descriptive or candidate-confusability shifts, since these settings increase distractor similarity without directly establishing a strict increase in minimum message complexity.}
While this phenomenon is widely recognized as a major failure mode in other domains, such as length generalization in sequence-to-sequence models \cite{lake2018generalization, anil2022exploring}, algorithmic extrapolation to larger data structures \cite{velickovic2019neural}, and visual reasoning with a larger number of objects than seen during training \cite{locatello2020object}, its implications for emergent communication protocols remain largely unexplored.

\begin{figure}[t]
  \includegraphics[width=\linewidth]{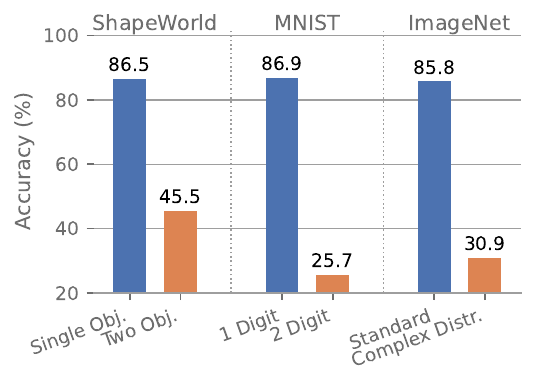}
  \caption{Accuracy (\%) of REINFORCE across datasets, showing performance differences between in-distribution and out-of-distribution test conditions.}
  \label{fig:reinforce}
\end{figure}

In referential games settings, the test-time environment can differ from training in at least two ways. 
First, the inputs may become more complex: for instance, agents might train on images featuring a single object but be evaluated on scenes with multiple objects (Fig.~\ref{fig:complexities}, right). 
\newrev{Second, the candidate set can contain more confusable distractors. During training, distractors may be easy to distinguish from the target. At test time, they may look similar to it, so a longer description may be needed (Fig.~\ref{fig:complexities}, left).}
In both cases, messages that were sufficient during training can become under-specified and ambiguous, causing the receiver to fail. 
\newrev{For example, ``dog'' may identify a target when the distractors are cars or airplanes. It becomes inadequate when the distractors are other breeds of dogs. In a multi-object setting, ``red square'' may identify a single-object training image. The same message cannot distinguish a target containing a red square and a blue circle from a distractor containing a red square and a green triangle.}
\newrev{Fig.~\ref{fig:reinforce} shows that REINFORCE~\cite{williams1992simple} loses accuracy as test batches become harder. On ShapeWorld~\cite{kuhnle2017shapeworld} and MNIST, performance drops when test images contain two objects or digits. On ImageNet, accuracy also declines when the candidate set contains visually similar images from the same class rather than randomly selected distractors.}

\newrev{To address this problem, the sender needs to assess whether its message distinguishes the target from the distractors before transmission. We propose Test-Time Adaptation (TTA) based on the self-play mechanism of~\cite{samiei2026vqel}. In that work, self-play grounds the sender's language during training on in-distribution (ID) data. Here, we reuse the resulting internal Message Perception Module as a verifier at test time. The sender uses this verifier to optimize its Message Generation Module before sending a message to the fixed receiver. This procedure requires no additional dialogue and can produce longer, more discriminative messages when the test setting demands them.}
\rev{It is worth noting that while utilizing a self-play verifier serves as a broad conceptual framework for TTA in emergent languages, in this work we implement and evaluate a specific formulation of this verifier based on the VQ-EL architecture. This demonstrates how specific architectural choices during training can be explicitly repurposed for test-time robustness.}

\begin{figure*}
  \includegraphics[width=\linewidth]{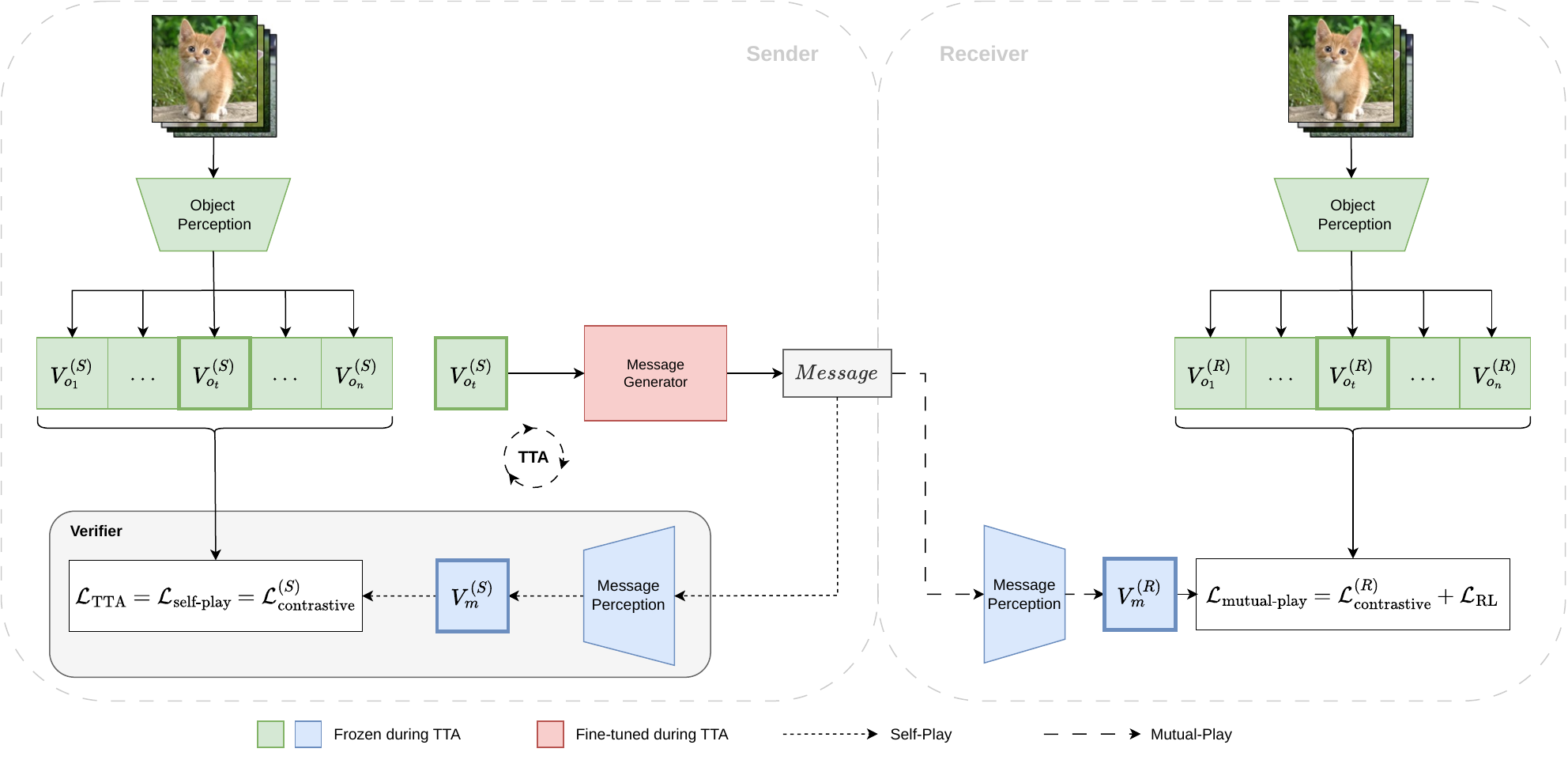}
  \caption{Overview of the self-play, mutual-play, and TTA framework. The sender first develops an internal language through self-play, and then communicates with a receiver during mutual play to establish a shared communication protocol. Self-play enables the sender to learn a \rev{message perception module}, which is subsequently reused as a verifier during TTA.}
  \label{fig:architecture}
\end{figure*}

In summary, we investigate how emergent referential protocols behave under test-time shifts that increase descriptive demands, and we propose a sender-side adaptation procedure that leverages self-play to refine messages prior to transmission. Our main contributions are threefold:
\begin{enumerate}
    \item An evaluation setting that tests emergent communication under increased test-time difficulty: richer inputs, more confusable distractors, and higher required solution complexity than in training.
    \item A test-time adaptation mechanism for emergent language, enabling the sender to adjust its communication strategy at deployment without additional interaction rounds with the receiver.
    \item Empirical validation showing that test-time adaptation improves robustness and communication success under these shifts.
\end{enumerate}

\section{Related Works}
\subsection{Emergent Language and Generalization}
The study of emergent communication primarily centers on referential games, where artificial agents must develop a shared protocol to achieve cooperative goals \cite{lazaridou2016multi, Havrylov2017EmergenceOL, lazaridou2020emergent, brandizzi2023toward}.
A substantial body of literature has explored how to shape these emergent protocols to be more robust, human-like, or structured.
Techniques include imposing architectural bottlenecks \cite{resnick2020capacity, rita2020lazimpa, eccles2019biases}, utilizing population-based training dynamics \cite{Gupta2021DynamicPM}, and applying iterated learning inspired by human cultural evolution \cite{Ren2020CompositionalLE}.
\rev{Despite their differences, these methods all intervene at \emph{training time} and leave the sender's policy frozen afterwards; none equips the agent to react when an individual test instance demands more communicative precision than training ever required. They are thus complementary to, rather than substitutes for, our deployment-time adaptation.}

A major focus within this domain is evaluating how well emergent languages generalize to unseen data.
\rev{A critical challenge in this setting is the agents' inherent bias to memorize and overfit to the training distribution, which severely hinders their ability to generalize to novel configurations \cite{rita2022emergent, bouchacourt-baroni-2018-agents}.}
Historically, this has been framed around \textit{compositional generalization}, the ability of agents to accurately describe novel combinations of previously seen attributes (e.g., a new combination of color and shape) \cite{chaabouni-etal-2020-compositionality, Kottur2017NaturalLD}.
\rev{Yet even the link between compositional structure and generalization is contested---noncompositional languages can generalize equally well in some tasks \cite{kharitonov2020generalization}, while compositionality appears essential under stricter evaluation protocols \cite{auersperger2022defending}---and the expressivity of an emergent language has been shown to be bounded by the contextual complexity experienced during training \cite{guo2021expressivity}, indicating that protocol structure alone does not guarantee robustness.}

\rev{More recently, several works have probed harder generalization regimes that require richer perception or reasoning, including communicating abstract rules over sets of objects \cite{mu2021emergent}, multi-object reference \cite{Feng2023MultiObject}, spatial and temporal relations \cite{Lipinski2024Spatial, Lipinski2023Time}, and rule-based reasoning \cite{Guo2023RulesReasoning}. Collectively, these studies demonstrate that protocols which appear successful in-distribution often collapse once the environment demands finer or relational distinctions. Their remedies, however, again operate at training time, enriching the training environments so that the required complexity is represented in the training distribution, and thus presuppose that the maximal test-time complexity is known in advance, an assumption our test-time adaptation explicitly removes.}

\newrev{In our work, the multi-object MNIST and ShapeWorld settings provide a direct operationalization of \textit{Complexity Out-of-Distribution (Complexity-OoD)} generalization \cite{samiei2026bridging}. We also study related object-descriptive and candidate-confusability shifts on ImageNet and COCO.}
\newrev{It is important to distinguish these settings from adjacent generalization paradigms. Standard distribution shifts typically involve visual or stylistic domain changes that leave the required solution complexity unchanged, and compositional generalization evaluates unseen \emph{combinations} of previously seen attributes while the number of attributes to be communicated remains constant. In the strict Complexity-OoD setting considered here, the minimal required solution complexity exceeds that of all training instances. The MNIST and ShapeWorld test images contain more objects than any training image, directly increasing the combinatorial descriptive demand. For ImageNet and COCO, we use the narrower terminology of object-descriptive or candidate-confusability shifts: these experiments establish more similar and difficult distractors, but do not directly measure a strict increase in the minimum message complexity.}

\subsection{Efficiency, Robustness, and Message Length in Communication}
A critical, yet under-explored, dimension of emergent protocols is the management of message length and communicative efficiency. Driven by optimization pressures, artificial agents frequently exhibit a ``principle of least effort,'' favoring the shortest possible messages that satisfy the training objective \cite{Evtimova2017MultiStep, kharitonov2020entropy, rita2020lazimpa}.
While researchers have studied the trade-offs between efficiency and informativeness, sometimes finding that artificial languages can be uniquely ``anti-efficient'' depending on the architecture \cite{Chaabouni2019AntiEfficient}, the inherent tendency to compress information becomes a severe liability under distributional shifts.
\rev{Most related to our architectural choices, Tucker et al. \cite{tucker2025vqvib} propose VQ-VIB, an information-constrained framework that, like ours, employs a vector-quantized discrete channel and explicitly controls the trade-off between informativeness and complexity. Crucially, however, this trade-off is fixed by training-time pressures: the achievable message complexity is calibrated to the training distribution and cannot be renegotiated when a test instance demands finer distinctions. Our framework is complementary; whereas existing methods tune the efficiency-informativeness balance during training, our sender re-optimizes it per instance at deployment.}
When test conditions become harder (e.g., highly confusable distractors or multi-object scenes), these brief messages become fundamentally under-specified and ambiguous.

\rev{A related line of work studies how channel noise shapes emergent protocols, ranging from its effect on message-length statistics \cite{ueda2021lewis} to the emergence of redundant encodings that act as an implicit repair mechanism \cite{implicitrepair2025}. These approaches improve robustness to \emph{channel-level} perturbations, but the resulting redundancy is learned at training time against an anticipated noise distribution. In contrast, we address \emph{instance-level} complexity shifts---where the message content itself, not its transmission, becomes insufficient---which no fixed redundancy scheme can resolve.}

To address communication failures at deployment, previous literature typically relies on multi-turn interactions. For instance, agents can be trained to issue clarification requests or engage in conversational repair \cite{Nikolaus2024Repair, hawkins2020continual}.
Other approaches address protocol misalignment by introducing intermediary translation modules or joiner agents \cite{Cope2024CLAP, Levy2025UnsupervisedTranslation}.
\rev{While multi-turn interactions and conversational repair mechanisms \cite{Nikolaus2024Repair, hawkins2020continual} are highly effective when both agents can continuously negotiate, they inherently demand multiple rounds of back-and-forth communication and a responsive receiver. In contrast, our proposed single-turn TTA operates entirely on the sender's side before transmission. This makes it a distinct and advantageous alternative for strict one-shot referential settings where communication bandwidth is limited, or where the receiver's policy is strictly frozen and cannot participate in repair dialogues.}

\subsection{Test-Time Training and Adaptation}
Test-Time Adaptation (TTA) encompasses a family of techniques designed to improve Out-of-Distribution (OoD) robustness by updating a deployed model (or its inputs) using only unlabeled test data \rev{\cite{liang2025ttasurvey, wang2024ottasurvey}}.
Early approaches in computer vision adapted batch-normalization statistics to match the target domain (e.g., AdaBN) \cite{Li2016RevisitingBN}.
More recent methods perform active optimization at inference time: Test-Time Training (TTT) updates network parameters by optimizing a self-supervised auxiliary loss concurrently on test inputs \cite{sun2020ttt}, while fully test-time adaptation methods like Tent minimize prediction entropy on the fly, typically updating only normalization-related parameters \cite{wang2020tent}.

Subsequent research has significantly expanded the scope and stability of TTA, addressing challenges such as single-sample adaptation, augmentation-guided updates \cite{Zhang2025TestTimeTD}, and model collapse under streaming or non-stationary data shifts \cite{wang2022cotta, niu2022eata, niu2023sar}.
\rev{In particular, stability under ``wild'' conditions---mixed shifts, small batches, and imbalanced label distributions---has been shown to hinge on which parameters are updated and how the self-supervised objective is regularized \cite{niu2023sar}, a concern that directly informs our decision to restrict adaptation to the message generator while freezing the perception modules. Another active direction is reducing the cost of adaptation itself, e.g., through backpropagation-free, forward-only optimization \cite{niu2024foa}; such techniques are complementary to our framework and offer a promising route to mitigating the inference-time overhead we report in Section~\ref{sec:limitations}. Notably, the standard taxonomy of TTA distinguishes between adapting to a single test batch and adapting to the overall test distribution \cite{liang2025ttasurvey}; our batch-level and dataset-level variants instantiate precisely these two regimes within emergent communication.}

\rev{Despite this breadth, existing TTA methods almost exclusively adapt \emph{continuous} predictors---image classifiers or dense regressors---using entropy minimization over a fixed label space or input-level self-supervision. None of these objectives transfers directly to our setting, where the adapted component is a \emph{discrete sequential message generator} and no label space or reconstruction target exists.}
Our proposed framework is heavily inspired by this broader TTA perspective, but we pioneer its application within the context of emergent communication. Rather than updating an image classifier, we adapt only the sender's discrete message generator at test time. By utilizing an internal self-supervised alignment objective \rev{---made available by the self-play architecture---} as the test-time loss, we enable instance-dependent message refinement under deployment-time shifts, entirely bypassing the need for receiver-side gradients.

\section{Preliminaries}
\subsection{Referential Game}

We study a one-shot referential game, a standard benchmark for emergent communication.
In each episode, the environment samples a \emph{candidate set} of $N$ objects
$\mathcal{O}=\{o_1,\dots,o_N\}$ and selects a target index $y \in \{1,\dots,N\}$.
The sender observes the target object $o_y$ among the candidate objects and produces a discrete message
$m=(w_1,\dots,w_L)$ of length $L$, where each token $w_t$ is chosen from a vocabulary
$\mathcal{V}$. The receiver observes the candidate set $\mathcal{O}$ and the message $m$,
and must predict the target index.

\subsection{Self-Play in Emergent Language}

We adopt a self-play based emergent communication framework. In the first stage, the sender engages in a referential game with itself, autonomously developing a communication protocol through \textbf{self-play}. In the second stage, the sender interacts with a receiver agent, refining and adapting the emergent language through \textbf{mutual-play}. 

Depending on their role (sender or receiver), agents are composed of a subset of the following three modules:
(i) an \textit{Object Perception Module} that encodes an input object $o$ into a vector representation $\mathbf{v}_o \in \mathbb{R}^d$, 
(ii) a \textit{Message Generation Module} that maps $\mathbf{v}_o$ to a sequence of discrete symbols through an RNN and a vector-quantized bottleneck (described in detail below), and 
(iii) a \textit{Message Perception Module} that encodes a message into a vector representation $\mathbf{v}_m$.

The message generator produces latent continuous states that are quantized using a learned codebook, yielding discrete units that function as symbols. During self-play, the model operates on the quantized embeddings, while in mutual-play it exchanges symbolic indices. The message perception module processes either representation and maps it back into the shared embedding space.

\paragraph{Self-Play}
In the self-play phase, a single agent plays a referential game with itself.

\rev{Given an object representation $\mathbf{v}_o$, the agent generates a discrete message, re-encodes it into $\mathbf{v}_m$, and optimizes an alignment objective. Specifically, given a batch of candidate objects $\mathcal{O}$ containing the target $o_y$, the self-play objective is formulated as a contrastive loss that maximizes the similarity between the message embedding and the target embedding while pushing away distractors:}
\rev{
\begin{equation}
\begin{aligned}
    \mathcal{L}_{\text{self-play}} 
    &= \mathcal{L}_{\text{contrastive}}\!\left(\mathbf{v}_m^{(S)}, \mathbf{v}_{o_y}^{(S)}\right) \\
    &= - \log \frac{\exp\left(\text{sim}(\mathbf{v}_m^{(S)}, \mathbf{v}_{o_y}^{(S)}) / \tau\right)}{\sum_{o' \in \mathcal{O}} \exp\left(\text{sim}(\mathbf{v}_m^{(S)}, \mathbf{v}_{o'}^{(S)}) / \tau\right)},
\end{aligned}
\end{equation}
}
This phase allows the agent to establish a structured discrete communication protocol before interacting with other agents.

\paragraph{Mutual-Play}
In the mutual-play phase, two agents assume the roles of sender and receiver. The sender encodes a target object and produces a sequence of symbolic tokens, which are transmitted to the receiver. The receiver encodes the message and selects the correct object among distractors using a similar alignment objective.

\rev{The receiver is trained via standard gradient-based optimization using the contrastive loss. Conversely, because the communication channel is discrete, the sender is updated using the REINFORCE policy-gradient algorithm. Instead of a sparse binary reward, the sender receives a continuous reward $R$ formulated as the negative of the receiver's cross-entropy loss (i.e., $R = -\mathcal{L}_{\text{contrastive}}$). The sender's RL objective is to maximize this expected reward, yielding the overall mutual-play loss:}
\rev{
\begin{equation}
    \mathcal{L}_{\text{mutual-play}} 
    = \mathcal{L}_{\text{contrastive}}\!\left(\mathbf{v}_m^{(R)}, \mathbf{v}_o^{(R)}\right) - \mathbb{E}\left[ R \log \pi_{\theta_{\text{MG}}^{(S)}}(m \mid \mathbf{v}_o^{(S)}) \right],
\end{equation}
where $\pi$ is the sender's message generation policy, parameterized by $\theta_{\text{MG}}^{(S)}$.}

Overall, training is sequential: the sender is first optimized in self-play to induce an initial discrete protocol grounded in object representations. It then enters mutual-play with a receiver, where this protocol is further adapted and refined through interactive communication.

\section{Test-Time Adaptation (TTA) in EL}
\subsection{Motivation of Sender-Side TTA}

\iftrue
\newrev{At deployment, the sender cannot access optimization signals from the receiver's Message Perception Module. Mutual-play has encouraged the two agents to develop compatible message representations. This motivates using the sender's own Message Perception Module as an internal verifier. In Section~\ref{sec:limitations}, CKA provides evidence of representational similarity, but not of identical decision boundaries or functional agreement under OoD conditions. The verifier is therefore an informative, imperfect proxy for the receiver.}

\newrev{During TTA, we update only the Message Generation (MG) Module. The sender's Object Perception (OP) and Message Perception (MP) Modules remain frozen. This choice helps prevent trivial changes to the representation space, including a possible collapse toward non-discriminative embeddings. Without ground-truth labels, jointly updating the perception modules could let the sender change both sides of the contrastive objective without improving the messages interpreted by the fixed receiver. Keeping these modules frozen instead makes them stable semantic anchors. The generator must then adapt its sequence of discrete symbols against those fixed representations. The full-sender ablation in Appendix~\ref{app:full-adaptation} shows that updating all sender modules leads to instability and representation drift.}

\newrev{The external receiver also remains fixed. The sender refines its message before transmission, preserving the one-shot setting without a receiver update, feedback round, or receiver-side gradient. This design isolates sender-side adaptation, but it has a limit: a receiver trained mainly on short messages may struggle with substantially longer ones under severe shifts. Lightweight receiver adaptation is a possible extension, though it would define a different setting.}

\newrev{The internal verifier may also diverge from the receiver under severe or unusual OoD shifts. In that case, a message that scores well internally may still be ambiguous to the receiver.}
\fi

\subsection{TTA Objective}
\label{sec:tta-objective}

In our framework, the Message Generation Module is implemented using an RNN and can therefore be unrolled for more time steps than the maximum length observed during training. This enables generating messages with length $L_{\text{test}} > L_{\text{train}}$.

However, directly increasing the length at test time without adaptation is typically ineffective. The generator has not been trained to produce coherent or informative tokens beyond $L_{\text{train}}$, and the additional tokens may become uninformative or even detrimental. Consequently, the Message Perception Module may fail to interpret the extended sequence meaningfully.

Our TTA mechanism resolves this limitation by utilizing the sender's internal self-play objective as an optimization signal directly at deployment. Concretely, during adaptation, we unroll the Message Generation Module for $L_{\text{test}}$ steps and optimize its parameters to minimize the self-play loss evaluated over the extended horizon:
\begin{equation}
    \mathcal{L}_{\text{TTA}}(L_{\text{test}}) =
    \mathcal{L}_{\text{self-play}}(\mathcal{B}_{\text{test}}; L_{\text{test}}).
\end{equation}
\newrev{This explicitly encourages the generator to select meaningful and discriminative tokens at later time steps, directly improving the alignment between the generated message and the target object under the sender's own Message Perception Module. In essence, the core objective of this adaptation is to adjust the Message Generation Module such that the resulting message effectively distinguishes the target from distractors from the sender's own internal perspective. Because mutual play promotes representational compatibility between the sender and receiver, sender-side discriminability provides a useful, though imperfect, proxy for messages that the fixed receiver can interpret.}

\newrev{To improve robustness to modest variations in message length, we randomly select the message length for each training batch from the set $\{L_{\text{train}} - 1, L_{\text{train}}\}$. This randomized length training (RLT) is a supporting training-time measure: our ablation in Appendix~\ref{app:random-length-trick} shows that it improves length robustness for all evaluated methods and provides an important preparation for extrapolation beyond the training length. It should therefore not be attributed to TTA alone. Crucially, RLT is applied identically to VQEL, VQEL+TTA, REINFORCE, and GS-ST in the principal comparisons. Consequently, the performance difference between vanilla VQEL and VQEL+TTA measures the additional contribution of test-time adaptation under matched training conditions.}

\subsection{Test-Time Optimization}

Let the sender, denoted by $S$, be parameterized by
\begin{equation}
    \theta^{(S)} = \{\theta_{\text{OP}}^{(S)}, \theta_{\text{MG}}^{(S)}, \theta_{\text{MP}}^{(S)}\},
\end{equation}
corresponding to the Object Perception Module, Message Generation Module, and Message Perception Module, respectively.

At test time, we freeze all sender modules except the Message Generation Module. That is, $\theta_{\text{OP}}^{(S)}$ and $\theta_{\text{MP}}^{(S)}$ are fixed, while the message generator's parameters are adapted. 

Given a test batch $\mathcal{B}_{\text{test}}$ (e.g., the current candidate set $\mathcal{O}$ if available to the sender, or a small buffer of test inputs),
\rev{we initialize the adapted weights as $\tilde{\theta}_{\text{MG}} \leftarrow \theta_{\text{MG}}^{(S)}$ and } perform $K$ gradient steps on:
\begin{equation}
\label{eq::tta-obj}
\mathcal{L}_{\text{TTA}}(\mathcal{B}_{\text{test}})=
\mathcal{L}_{\text{self-play}}(\mathcal{B}_{\text{test}}).
\end{equation}
The inner-loop update is:
\begin{equation}
\tilde{\theta}_{\text{MG}} \leftarrow
\tilde{\theta}_{\text{MG}} - \eta
\nabla_{\tilde{\theta}_{\text{MG}}} \mathcal{L}_{\text{TTA}},
\end{equation}
where $\eta$ is the adaptation learning rate.

After adaptation, the sender generates the final symbolic message using the adapted Message Generation Module:
\begin{equation}
m^{*} =
f_{\text{MG}}^{(S)}\!\left(
f_{\text{OP}}^{(S)}(o_y);
\tilde{\theta}_{\text{MG}}
\right).
\end{equation}

\rev{The complete step-by-step procedure for this test-time optimization is summarized in Algorithm \ref{alg:tta}.}

\newrev{A stopping criterion determines when to end the test-time optimization loop. In other TTA settings, adaptation may use a single update per batch or continue until an auxiliary loss converges. Our experiments instead use a fixed budget of $K$ gradient steps per test instance, as reported in Section~\ref{sec:limitations}. This provides a predetermined stopping point and bounds the number of updates. A dynamic alternative would stop early if the internal self-play loss $\mathcal{L}_{\text{TTA}}$ plateaus, suggesting that further updates yield little improvement under the frozen verifier. Such a rule might reduce computational overhead. We have not evaluated it, however, and do not claim that it would preserve communication accuracy. Its accuracy--latency trade-off, like that of other lower-cost variants, remains to be established experimentally.}

\subsection{TTA at Different Data Availability Levels}

Our TTA objective can be instantiated at different levels depending on what is available to the sender at deployment.

\paragraph*{Batch-level TTA}
If the sender has access to a batch containing the target and distractors (e.g., the candidate set of the referential game), TTA can be performed using the full contrastive self-play objective. Let $\mathcal{B}=\{o_1,\dots,o_B\}$ be the batch and $o_i$ the target. We minimize $\mathcal{L}_{\text{TTA}}(\mathcal{B})$ as defined in Eq.~\ref{eq::tta-obj}.

Batch-level TTA explicitly accounts for the current distractor distribution and encourages messages that are maximally discriminative with respect to the available negatives.
\rev{It is important to acknowledge that this formulation grants the sender an informational advantage at inference time—namely, explicit access to the specific distractor distribution—which the standard non-adaptive baselines do not utilize.}
\newrev{This access is also part of the intended adaptation setting: the contribution of TTA is to convert the available candidate-set structure into a self-supervised contrastive signal through the sender's internal verifier. The standard REINFORCE and GS-ST baselines, in their fixed formulations, do not contain an equivalent grounded verifier or test-time objective through which they could directly exploit this signal. Adding such a mechanism would require a separately designed adaptive variant rather than the original fixed baseline.}

\paragraph*{Dataset-level TTA.}
If the sender has access to the entire unlabeled test dataset $\mathcal{D}_{\text{test}}$ (or a sufficiently large stream), we minimize the expected batch-level objective:
\begin{equation}
\mathcal{L}_{\text{TTA-dataset}}
=
\mathbb{E}_{\mathcal{B} \sim \mathcal{D}_{\text{test}}}
\left[
\mathcal{L}_{\text{TTA}}(\mathcal{B})
\right].
\end{equation}

This variant adapts the sender to the overall statistics of the test distribution and can be interpreted as test-time domain adaptation for emergent communication, while restricting updates to the Message Generation Module parameters.
\rev{Specifically, batches of candidate sets are uniformly sampled from the unlabeled test dataset. For each test instance, the sender observes the assigned target object and minimizes the self-play loss using the available distractors in the sampled batch.}
\newrev{Dataset-level TTA removes the episode-specific advantage of adapting to the exact distractors in each evaluation episode, but it still assumes access to an unlabeled test stream and performs additional gradient-based optimization. These information and computation requirements are therefore part of the deployment assumptions of the proposed adaptation setting.}

\begin{algorithm}[t]
\caption{{Sender-Side Test-Time Adaptation (Batch-Level)}}
\label{alg:tta} 
\begin{algorithmic}[1]
\STATE \textbf{Input:} Pre-trained Sender $S$ parameterized by $\theta_{\text{OP}}^{(S)}, \theta_{\text{MG}}^{(S)}, \theta_{\text{MP}}^{(S)}$
\STATE \textbf{Input:} Test batch $\mathcal{B}_{\text{test}} = \{o_1, \dots, o_B\}$ with target $o_y$
\STATE \textbf{Input:} Number of adaptation steps $K$, learning rate $\eta$
\STATE \textbf{Initialize:} Freeze $\theta_{\text{OP}}^{(S)}$ and $\theta_{\text{MP}}^{(S)}$
\STATE \textbf{Initialize:} $\tilde{\theta}_{\text{MG}} \leftarrow \theta_{\text{MG}}^{(S)}$ \COMMENT{Copy parameters for test instance}
\FOR{$k = 1, \dots, K$}
    \STATE Compute object embeddings: $\mathbf{v}_{o}^{(S)} = f_{\text{OP}}^{(S)}(o; \theta_{\text{OP}}^{(S)})$ for $o \in \mathcal{B}_{\text{test}}$
    \STATE Generate extended messages: $m = f_{\text{MG}}^{(S)}(\mathbf{v}_{o}^{(S)}; \tilde{\theta}_{\text{MG}})$
    \STATE Compute message embeddings: $\mathbf{v}_{m}^{(S)} = f_{\text{MP}}^{(S)}(m; \theta_{\text{MP}}^{(S)})$
    \STATE Compute TTA loss: 
    $$\mathcal{L}_{\text{TTA}} = \mathcal{L}_{\text{contrastive}}\!\left(\mathbf{v}_m^{(S)}, \mathbf{v}_o^{(S)}\right)$$
    \STATE Update generator: 
    $$\tilde{\theta}_{\text{MG}} \leftarrow \tilde{\theta}_{\text{MG}} - \eta \nabla_{\tilde{\theta}_{\text{MG}}} \mathcal{L}_{\text{TTA}}$$
\ENDFOR
\STATE \textbf{Output:} Adapted message $m^{*} = f_{\text{MG}}^{(S)}\!\left(f_{\text{OP}}^{(S)}(o_y); \tilde{\theta}_{\text{MG}}\right)$
\end{algorithmic}
\end{algorithm}

\section{Experimental Setup}
\subsection{Datasets}

We evaluate our method on \newrev{five} datasets. Below, we briefly describe each dataset along with its corresponding ID and OoD splits. \newrev{Regarding the nature of the OoD shifts, MNIST, ShapeWorld, and Visual Genome correspond to multi-object Complexity-OoD, while ImageNet and COCO correspond to object-descriptive or candidate-confusability shifts.} Refer to Appendix \ref{app:datasets} for sample images from each dataset.

\subsubsection{MNIST}
MNIST \cite{lecun1998mnist} is a widely used dataset of handwritten digits ranging from 0 to 9, consisting of 60,000 training and 10,000 test images. Each image is grayscale and fixed size of $28 \times 28$ pixels.

\paragraph{1Digit} 
The 1Digit subset contains images with a single digit positioned either on the left or right side of the frame. This represents the  ID condition.

\paragraph{2Digit} 
To create the 2Digit subset, two images from the original MNIST dataset are combined into a single image, placing one digit on the left and one on the right. This simulates an OoD condition with increased number of objects. Refer to Figure~\ref{fig:mnist_samples} for an example.

\subsubsection{ShapeWorld}
ShapeWorld \cite{kuhnle2017shapeworld} is a synthetic visual dataset designed to test relational reasoning and compositional generalization. Images depict simple colored geometric shapes arranged on a grid. We control the number of objects per image to create different ID and OoD conditions.

\paragraph{Single-Object}
Each image contains exactly one object, representing the ID condition where the task is simpler and the referential target is unambiguous.

\paragraph{Two-Object}
Each image contains exactly two objects. The presence of multiple objects increases combinatorial complexity, creating an OoD evaluation scenario.

\subsubsection{ImageNet}
We also evaluate on the ImageNet-1k dataset \cite{deng2009imagenet} to test generalization to high-dimensional, naturalistic visual embeddings. 

\paragraph{Standard}
Each batch of samples is drawn randomly from the dataset. Consequently, some samples may belong to the same class, while others do not, providing a representative ID evaluation.

\paragraph{Complex}
For the OoD subset, all samples within a batch are drawn from the same class. This setup increases difficulty because the objects are close in the embedding space, requiring more precise messaging for accurate identification.

\rev{
\subsubsection{\rev{COCO}}
The COCO dataset \cite{lin2014microsoft} contains natural images with multiple objects and diverse scene contexts. Compared with the other datasets considered in this work, COCO represents a more challenging benchmark due to its greater visual complexity and scene variability.

\paragraph{\rev{Standard}}
Each batch is formed by randomly sampling images from the dataset. Consequently, images within a batch may contain semantically unrelated objects and scenes, providing a representative ID evaluation setting.

\paragraph{\rev{Complex}}
For the OoD subset, batches are constructed by grouping images based on nearest-neighbor relationships in the object perception embedding space. As a result, images within a batch are semantically similar, making candidate images more difficult to distinguish. This setting requires more precise and informative messages for accurate identification.
}

\subsubsection{\newrev{Visual Genome}}
\newrev{The Visual Genome dataset \cite{krishna2017visual} is used to evaluate generalization in a more naturalistic visual environment with diverse objects, relationships, and messier real-world contexts. Details of the dataset splits and evaluation results are provided in Appendix~\ref{app:genome}.}

\subsection{Baselines}
\label{sec:baselines}

We compare our TTA method against two commonly used baselines for emergent communication in referential games: REINFORCE \cite{williams1992simple} and GS \cite{jang2017categorical, maddison2017concrete}. To maintain backwards differentiability in GS, we use the straight-through (ST) trick \rev{\cite{jang2017categorical}} with learning the inverse-temperature  with a multilayer perceptron \rev{\cite{jang2017categorical}}:
\begin{equation}
    \frac{1}{\tau(h_i)} = \log (1 + \exp(\mathbf{w}_{\tau}^T \mathbf{h}_i)) + \tau_0
\end{equation}
where $\tau_0$ controls maximum possible value for the temperature \rev{and $h_i$ denotes the hidden state of the Message Generation Module's GRU at step $i$.}

\rev{
We also compare against an oracle baseline that serves as an upper bound. This baseline updates the same components as TTA using the receiver loss under the same computational budget and employs GS-ST for both ID training and OoD fine-tuning. An alternative oracle variant is evaluated in Appendix~\ref{app:oracles}.
}

\subsection{Agents' Architecture}

All three methods (VQEL, REINFORCE, and GS-ST) share the same overall architecture, with an identical number of parameters.

\paragraph{Object Perception Module}
\rev{
For ShapeWorld and MNIST, we use the CNN encoder from Prototypical Networks \cite{snell2017prototypical}. For ImageNet, we employ a ResNet-50 \cite{he2016deep} pretrained on ImageNet-1K, with all parameters frozen during training. For COCO, we use the pretrained DINOv2-Base encoder \cite{oquab2023dinov2}, keeping its parameters fixed throughout training.
}

\paragraph{Message Generation Module}  
This module comprises a GRU and a VQ module, with cosine similarity used as the distance metric in the VQ module. Note that GS-ST and REINFORCE do not include the VQ module.

\paragraph{Message Perception Module}  
It consists of a GRU and an embedding layer, which converts symbolic messages into embeddings before passing them through the GRU.

\subsection{Hyperparameter Settings}

\paragraph{Data Splits}
For all experiments, the ID dataset is partitioned into training, validation, and test splits with proportions of 80\%, 10\%, and 10\%, respectively. The OoD test split is constructed to match the size of the ID test set, ensuring a balanced evaluation protocol across conditions.

\paragraph{Communication Setup}
Across all methods, the vocabulary size is fixed to 10. During training, the message length is set to $L_{Train} = 4$. At test time, we evaluate models with message lengths ranging from 4 to 15 to assess generalization beyond the training configuration. Agents are trained with a batch size of 32 (corresponding to 31 distractors per target instance). Unless otherwise stated, evaluation on both ID and OoD subsets is conducted with a batch size of 100\newrev{ (corresponding to 99 distractors per target instance)}. To further analyze robustness, we additionally report performance under varying batch sizes.
\newrev{We deliberately use the larger evaluation batch to make the referential task more challenging, so that performance differences among the methods are easier to observe. The default evaluation batch size is the same for all methods.}
\newrev{Furthermore, as discussed in Section~\ref{sec:tta-objective}, the randomized training message-length strategy ($L \in \{3, 4\}$) is applied consistently to all evaluated methods, including the REINFORCE and GS-ST baselines. This shared protocol controls for the contribution of RLT when comparing the additional effect of test-time adaptation.}

\paragraph{Training and Optimization}
All models are trained using the Adam optimizer with a weight decay of $1 \times 10^{-5}$. The learning rate is selected via grid search over the range $[10^{-5}, 10^{-3}]$ on a logarithmic scale. The sampling temperature is tuned within $[10^{-6}, 1]$ with a step size of 0.1, and the initial temperature parameter of the GS-ST, $\tau_0$, is selected from $[0.1, 1.5]$ with a step size of 0.1.
\newrev{All reported quantitative results are averaged over three independent runs with different random seeds. Unless otherwise noted, tables report the mean $\pm$ standard deviation, and plots show the corresponding mean curves together with error bars or shaded variability regions representing one standard deviation across runs.}

Baseline methods are trained for 100 epochs. In contrast, VQEL is trained 50 epochs of self-play followed by 50 epochs of mutual play, yielding a total of 100 training epochs for fair comparison. \rev{We note that vocabulary size, codebook size, and training duration are training-time hyperparameters of the underlying VQEL framework; we keep them fixed and identical across all methods (vocabulary size 10, a matched codebook, and 100 total training epochs) so that the comparison isolates the effect of test-time adaptation rather than these shared settings.}

\paragraph{Test-Time Adaptation}
For TTA, we tune the adaptation learning rate over the set $\{10^{-4}, 10^{-5}\}$. For batch-level adaptation, we search over $\{50, 100, 200, 300\}$ update iterations. For dataset-level adaptation, the number of update iterations is tuned over $\{5, 10, 15, 20\}$. 

\rev{We select the optimal configuration using an elbow-based criterion applied to the verifier scores. Specifically, for each learning rate, we identify the corresponding optimal number of steps via the elbow method on the verifier score curve. We then compare these optimal configurations across learning rates and choose the learning rate that achieves the highest verifier score.}

\section{Results}
\subsection{Evaluation of baselines on ID and complex OoD}

\newrev{Tables \ref{tab:multi-object-ood} and \ref{tab:object-descriptive-ood} report the accuracy of GS-ST, REINFORCE, and VQEL on both the ID and OoD splits.} As shown, all three methods experience a substantial drop in performance when either multi-object complexity (MNIST and ShapeWorld) or object-descriptive/candidate-confusability shifts (ImageNet and \rev{COCO}) increase at test time. In many cases, the accuracy decreases by roughly half. The results indicate that none of the evaluated methods exhibit robustness to these distribution shifts.

\begin{table*}[t]
\centering
\setlength{\tabcolsep}{7pt}
\renewcommand{\arraystretch}{1.1}
\rev{
\begin{tabular}{l c c c c}
\toprule
\multirow{2}{*}{\textbf{Method}} 
& \multicolumn{2}{c}{\textbf{MNIST}} 
& \multicolumn{2}{c}{\textbf{ShapeWorld}} \\
\cmidrule(lr){2-3} \cmidrule(lr){4-5}
& 1Digit (ID) & 2Digit (OoD)
& Single-object (ID) & Two-object (OoD) \\
\midrule
GS-ST     & $80.2_{\pm 0.7}$ & $23.7_{\pm 3.3}$ & $84.2_{\pm 0.8}$ & $38.8_{\pm 0.7}$ \\
REINFORCE & $86.9_{\pm 2.3}$ & $25.7_{\pm 5.0}$ & $86.5_{\pm 0.2}$ & $45.5_{\pm 2.9}$ \\
VQEL      & $90.1_{\pm 1.5}$ & $42.0_{\pm 2.8}$ & $88.7_{\pm 0.2}$ & $47.9_{\pm 1.1}$ \\
\bottomrule
\end{tabular}
}
\caption{\rev{Test accuracy (\%) of the evaluated methods on the In-Distribution (ID) and Out-of-Distribution (OoD) splits for datasets characterized by multi-object complexity. For both MNIST and ShapeWorld, the OoD condition presents test instances with more objects than encountered during training, thereby increasing the combinatorial complexity. The substantial drop in accuracy across all standard baselines highlights the inherent vulnerability of fixed emergent protocols to multi-object complexity shifts.}}
\label{tab:multi-object-ood}
\end{table*}

\begin{table*}[t]
\centering
\setlength{\tabcolsep}{7pt}
\renewcommand{\arraystretch}{1.1}
\rev{
\begin{tabular}{l c c c c}
\toprule
\multirow{2}{*}{\textbf{Method}} 
& \multicolumn{2}{c}{\textbf{ImageNet}} 
& \multicolumn{2}{c}{\textbf{COCO}} \\
\cmidrule(lr){2-3} \cmidrule(lr){4-5}
& Standard (ID) & Complex (OoD)
& Standard (ID) & Complex (OoD) \\
\midrule
GS-ST     & $73.5_{\pm 0.9}$ & $35.4_{\pm 1.4}$ & $86.0_{\pm 0.5}$ & $40.9_{\pm 0.4}$ \\
REINFORCE & $85.8_{\pm 1.2}$ & $30.9_{\pm 1.3}$ & $88.8_{\pm 0.4}$ & $42.9_{\pm 1.3}$ \\
VQEL      & $89.7_{\pm 0.5}$ & $42.7_{\pm 2.1}$ & $89.3_{\pm 0.9}$ & $49.7_{\pm 1.3}$ \\
\bottomrule
\end{tabular}
}
\caption{\newrev{Test accuracy (\%) of the evaluated methods on the ID and OoD splits for datasets characterized by object-descriptive or candidate-confusability shifts. In ImageNet and COCO, the OoD scenarios introduce highly confusable distractors (e.g., candidates sharing the same visual class or multiple attributes), which demand finer and more specific message descriptions to accurately identify the target. Similar to the multi-object setting, all non-adaptive methods suffer a severe performance collapse under this distributional shift.}}
\label{tab:object-descriptive-ood}
\end{table*}

\subsection{Effect of TTA on Increasing Multi-Object Complexity}

\begin{figure*}[h]
    \centering
    \begin{minipage}{0.48\textwidth}
        \centering
        \includegraphics[width=\linewidth]{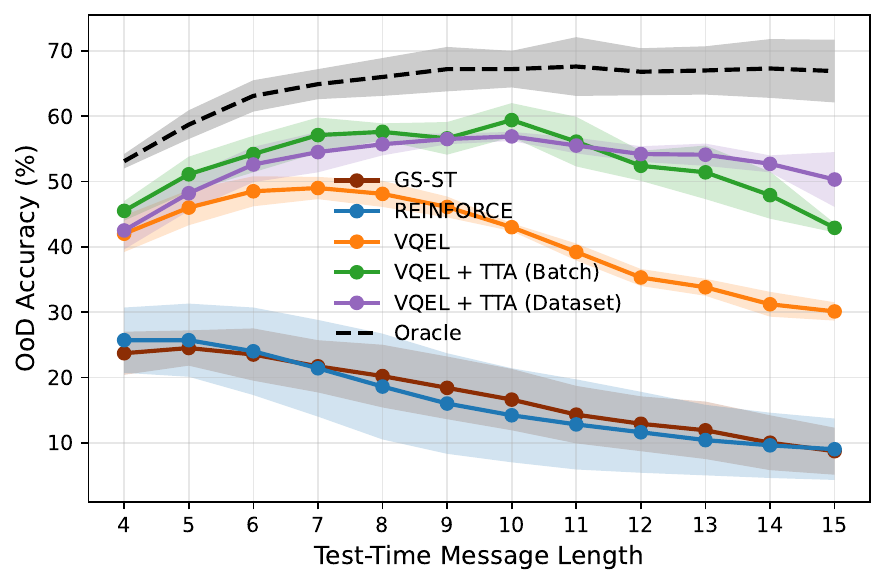}
    \end{minipage}
    \hfill
    \begin{minipage}{0.48\textwidth}
        \centering
        \includegraphics[width=\linewidth]{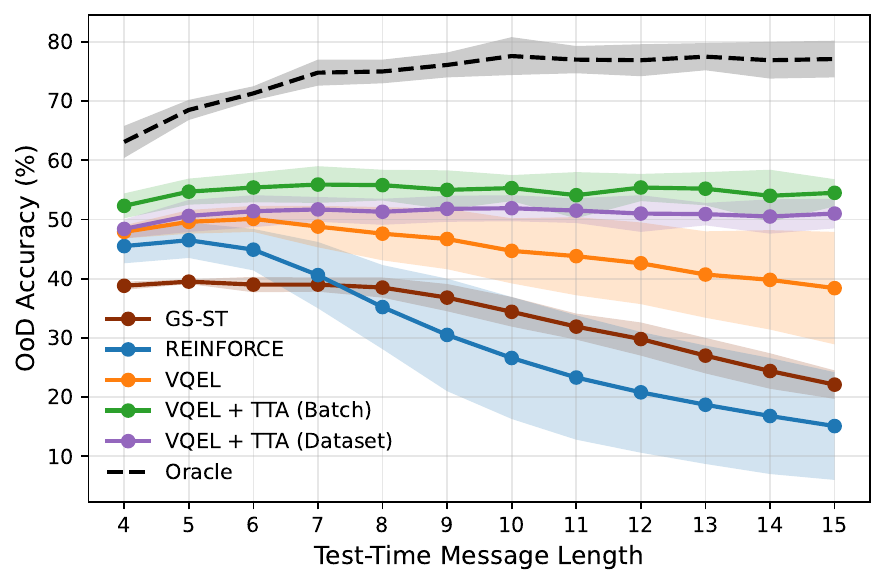}
    \end{minipage}
    \caption{\newrev{TTA accuracy on the OoD split. Curves show mean accuracy over three independent random seeds; shaded regions indicate one standard deviation. Left: MNIST dataset; right: ShapeWorld dataset.}}
    \label{fig:multi-object-complexity}
\end{figure*}

Figure \ref{fig:multi-object-complexity} reports the accuracy of TTA on the MNIST and ShapeWorld datasets as multi-object complexity increases. As previously described, we evaluated all methods with varying message lengths at test time. The accuracy improvement of TTA over baseline methods reached up to 35\% on MNIST. On \rev{ShapeWorld}, TTA achieved an accuracy 10\% higher than REINFORCE and 15\% higher than
GS-ST.

The results indicate that longer messages consistently improve TTA accuracy across both MNIST and ShapeWorld, with gains reaching up to 15\% on MNIST. This suggests that increased multi-object complexity necessitates longer messages for accurate target description. In contrast the baseline methods do not generalize effectively to longer message lengths, with their performance either plateauing or decreasing.

\newrev{The inclusion of the vanilla VQEL baseline (without test-time adaptation) in Figures \ref{fig:multi-object-complexity} and \ref{fig:object-descriptive-complexity} isolates the effect of TTA from the inherent architectural advantages of VQEL. All methods in these comparisons are trained with the same randomized length training protocol. RLT improves length robustness for every method, as shown in Appendix~\ref{app:random-length-trick}; it is therefore a supporting condition for extrapolation to longer messages rather than a benefit attributable only to TTA. Under this matched training protocol, vanilla VQEL still plateaus or degrades as test-time message length increases, whereas VQEL+TTA better exploits longer messages. The resulting gap thus quantifies the additional contribution of test-time adaptation beyond the shared training-time robustness measure.}

\newrev{Table~\ref{tab:component-summary} provides a concise summary of the component-wise evidence available in the current experiments. The VQEL versus VQEL+TTA comparison isolates the additional contribution of test-time optimization under the same VQEL architecture and randomized-length training protocol. Across all four datasets, the batch-level TTA variant improves over vanilla VQEL, with the largest absolute gains occurring on the more confusable ImageNet and COCO settings. The remaining components are supported by complementary ablations: the RLT ablation evaluates the training-time length-robustness contribution, the generator-only versus full-sender comparison evaluates the frozen-perception design, and the CKA analysis evaluates representational similarity between the internal verifier and receiver. These diagnostics support the respective roles of the components, but do not constitute independent causal ablations of every component in isolation.}

\begin{table*}[t]
\centering
\small
\setlength{\tabcolsep}{8pt}
\renewcommand{\arraystretch}{1.15}
\newrev{
\begin{tabular}{l c c c}
\toprule
\textbf{Dataset} & \textbf{VQEL} & \textbf{VQEL+TTA (Batch)} & \textbf{Absolute gain} \\
\midrule
MNIST      & $48.1_{\pm 2.0}$ & $59.4_{\pm 2.6}$ & $+11.3$ \\
ShapeWorld & $47.6_{\pm 4.5}$ & $55.9_{\pm 3.1}$ & $+8.3$ \\
ImageNet   & $43.9_{\pm 0.8}$ & $71.1_{\pm 2.1}$ & $+27.2$ \\
COCO       & $50.4_{\pm 1.6}$ & $66.5_{\pm 0.9}$ & $+16.1$ \\
\bottomrule
\end{tabular}
}
\caption{\newrev{Summary of the additional contribution of batch-level TTA over vanilla VQEL, using the setting-level message lengths and accuracies reported in Table~\ref{tab:max-accuracies}. The two methods share the VQEL architecture and randomized-length training protocol; therefore, the differences summarize the added effect of test-time optimization under matched training conditions.}}
\label{tab:component-summary}
\end{table*}

\newrev{On ShapeWorld, batch-level TTA increases mean OoD accuracy from $47.6\pm4.5\%$ for VQEL to $55.9\pm3.1\%$ at the setting-level selected message lengths. The $8.3$ percentage-point gain is smaller than the corresponding gains on the other three datasets, while remaining a notable increase in the observed mean.}

\subsection{Effect of TTA on Increasing Object Descriptive Complexity}

\begin{figure*}
    \centering
    \begin{minipage}{0.48\textwidth}
        \centering
        \includegraphics[width=\linewidth]{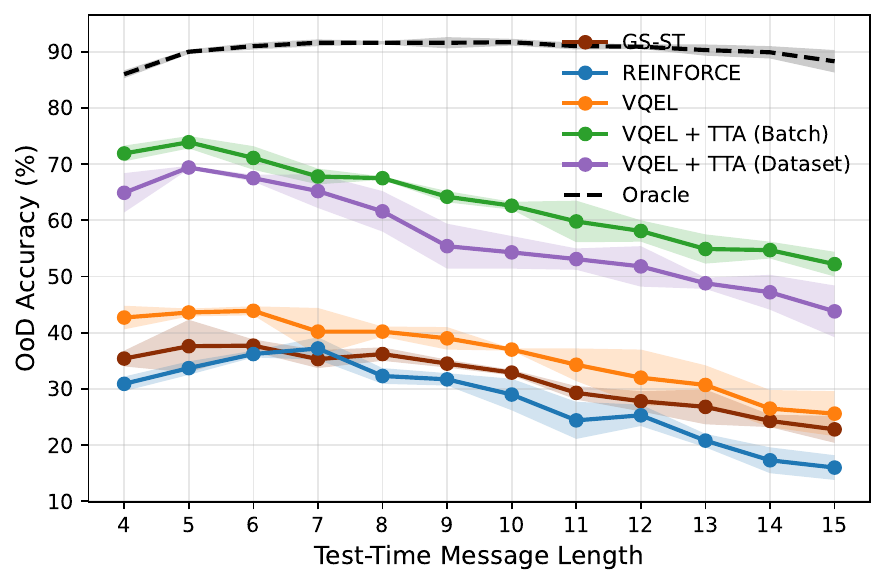}
    \end{minipage}
    \hfill
    \begin{minipage}{0.48\textwidth}
        \centering
        \includegraphics[width=\linewidth]{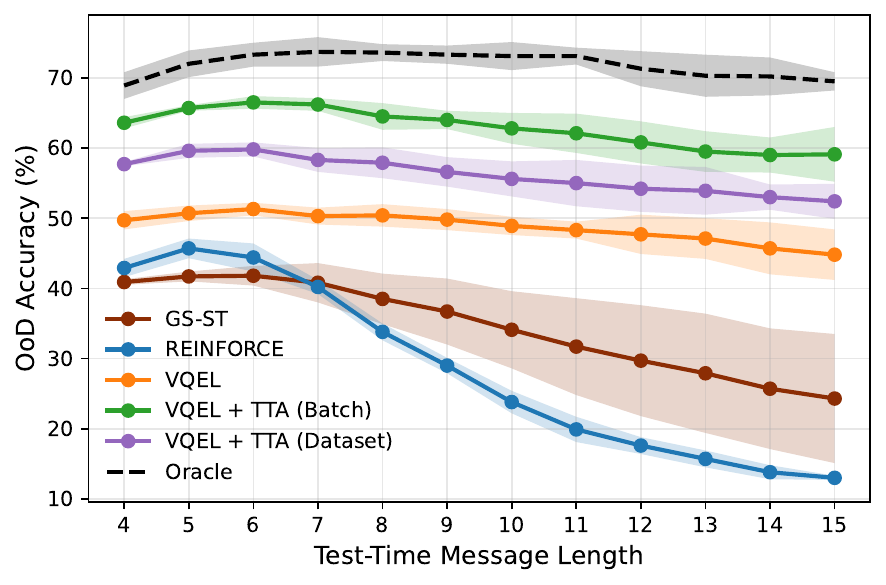}
    \end{minipage}
    \caption{\newrev{TTA accuracy on the OoD split. Curves show mean accuracy over three independent random seeds; shaded regions indicate one standard deviation. Left: ImageNet dataset; right: COCO dataset.}}
    \label{fig:object-descriptive-complexity}
\end{figure*}

Figure \ref{fig:object-descriptive-complexity} illustrates the OoD accuracy for the ImageNet and \rev{COCO} datasets. \rev{On the ImageNet dataset, TTA outperforms the baselines by a margin of 30\%,  and on COCO by about 16–20\%. On both datasets, adaptation enables the model to benefit from longer messages, with the elbow-selected length exceeding the training length, whereas the non-adaptive baselines plateau or degrade as message length grows.}
\newrev{As a complementary test-time strategy, sampling-based test-time scaling (TTS) yields consistent but comparatively modest OoD accuracy gains of approximately 2--6\% in our experiments; its sampling procedure and complete results on all four datasets are reported in Appendix~\ref{sec:appendix-b}.}

\subsection{\newrev{Setting-Level Message Length Selection}}
\newrev{We evaluate candidate message lengths over the full OoD split and select one length for each dataset--method setting. This length is then used for all test instances in that setting; it is not selected separately for each instance or batch. For VQEL and the TTA variants, we apply an elbow-based criterion to the sender-side self-play or verifier scores aggregated across the OoD split. REINFORCE and GS-ST lack an internal verifier, so we use their mutual-play accuracies as an oracle reference for length selection. Table~\ref{tab:max-accuracies} compares the methods at their selected setting-level lengths and reports the corresponding mean OoD accuracy.}

\begin{table*}
\centering
\small
\setlength{\tabcolsep}{6pt}
\renewcommand{\arraystretch}{1.15}
\rev{
\begin{tabular}{l l c c}
\toprule
\textbf{Dataset} & \textbf{Method} & \textbf{TT Msg. Len} 
& \textbf{Accuracy} \\
\midrule

\multirow{5}{*}{\textbf{MNIST}} 
& GS-ST                     & $5$           & $24.5_{\pm 2.7}$          \\
& REINFORCE                 & $4$           & $25.7_{\pm 5.0}$          \\
& VQEL                      & $8$           & $48.1_{\pm 2.0}$          \\
& \textbf{VQEL+TTA (Batch)} & $\mathbf{10}$ & $\mathbf{59.4_{\pm 2.6}}$ \\
& VQEL+TTA (Dataset)        & $10$          & $56.9_{\pm 0.7}$          \\

\midrule
\multirow{5}{*}{\textbf{ShapeWorld}} 
& GS-ST                     & $5$          & $39.5_{\pm 0.4}$          \\
& REINFORCE                 & $5$          & $46.5_{\pm 3.0}$          \\
& VQEL                      & $8$          & $47.6_{\pm 4.5}$          \\
& \textbf{VQEL+TTA (Batch)} & $\mathbf{7}$ & $\mathbf{55.9_{\pm 3.1}}$ \\
& VQEL+TTA (Dataset)        & $8$          & $51.3_{\pm 2.2}$          \\

\midrule
\multirow{5}{*}{\textbf{ImageNet}} 
& GS-ST                     & $6$          & $37.7_{\pm 1.1}$          \\
& REINFORCE                 & $7$          & $37.2_{\pm 2.0}$          \\
& VQEL                      & $6$          & $43.9_{\pm 0.8}$          \\
& \textbf{VQEL+TTA (Batch)} & $\mathbf{6}$ & $\mathbf{71.1_{\pm 2.1}}$ \\
& VQEL+TTA (Dataset)        & $5$          & $69.4_{\pm 0.3}$          \\

\midrule
\multirow{5}{*}{\textbf{COCO}} 
& GS-ST                     & $6$          & $41.8_{\pm 1.4}$          \\
& REINFORCE                 & $5$          & $45.7_{\pm 1.4}$          \\
& VQEL                      & $8$          & $50.4_{\pm 1.6}$          \\
& \textbf{VQEL+TTA (Batch)} & $\mathbf{6}$ & $\mathbf{66.5_{\pm 0.9}}$ \\
& VQEL+TTA (Dataset)        & $8$          & $57.9_{\pm 2.2}$          \\

\bottomrule
\end{tabular}
}
\caption{\newrev{Selected message length and corresponding mean OoD accuracy for each dataset--method setting. Scores are aggregated over the full OoD split to select one length per setting, not per test instance or batch. VQEL and TTA use an elbow-based criterion on sender-side self-play or verifier scores; REINFORCE and GS-ST use mutual-play accuracy as an oracle reference.}}
\label{tab:max-accuracies}
\end{table*}

\subsection{\rev{Qualitative analysis of the learned language}}

\begin{figure*}
    \centering

    \begin{minipage}{0.33\textwidth}
        \centering
        \includegraphics[width=\linewidth]{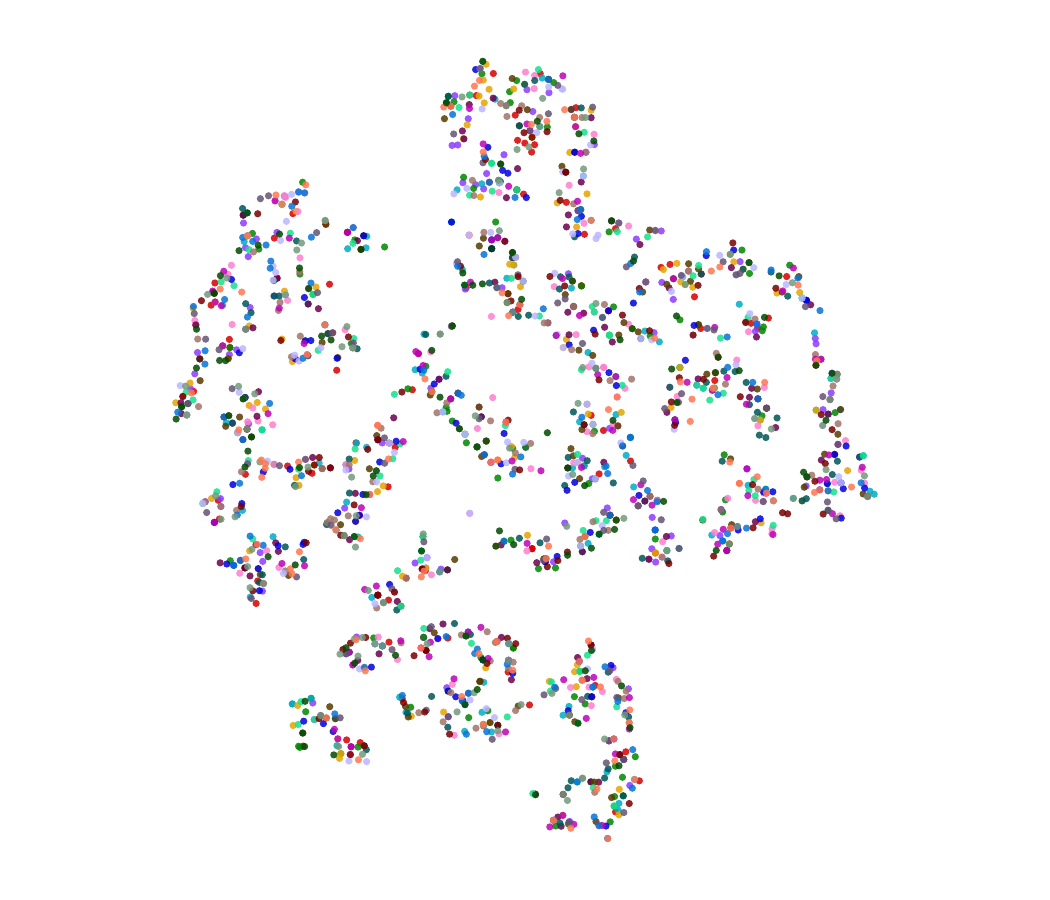}
        \\[2pt] (a)
    \end{minipage}
    \hfill
    \begin{minipage}{0.33\textwidth}
        \centering
        \includegraphics[width=\linewidth]{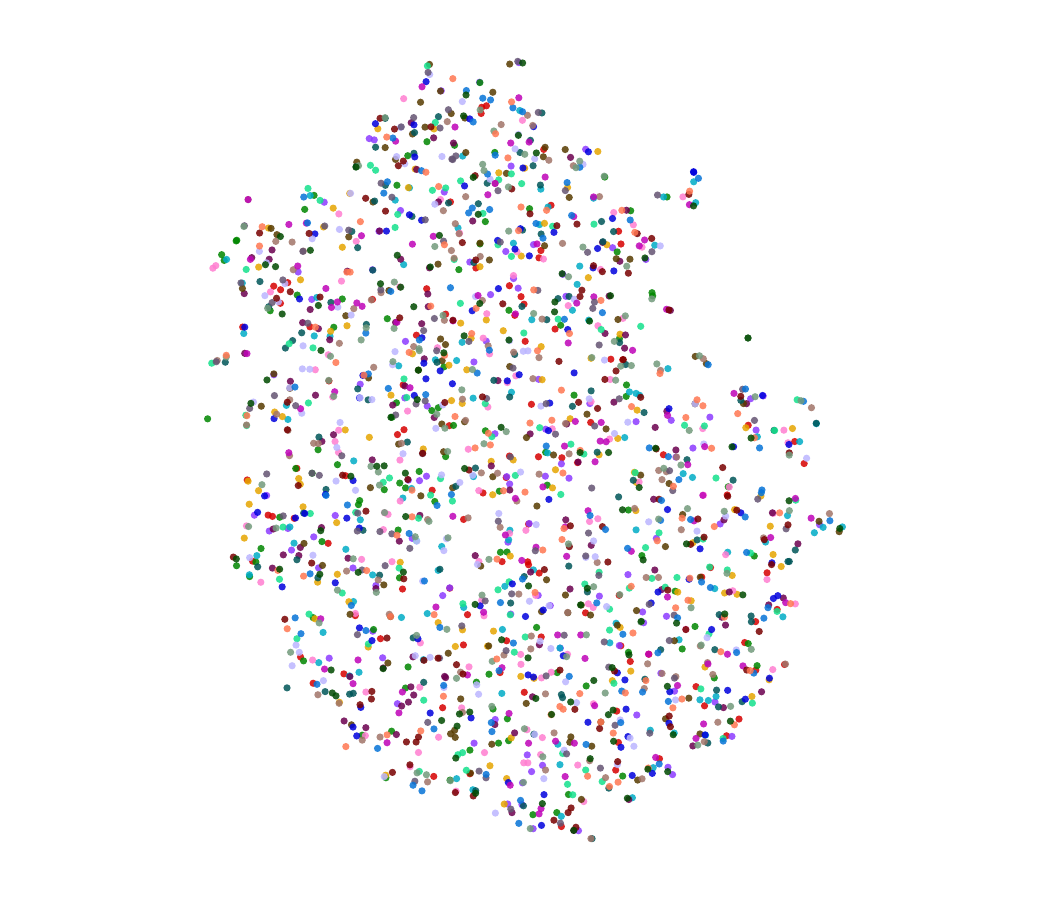}
        \\[2pt] (b)
    \end{minipage}
    \hfill
    \begin{minipage}{0.33\textwidth}
        \centering
        \includegraphics[width=\linewidth]{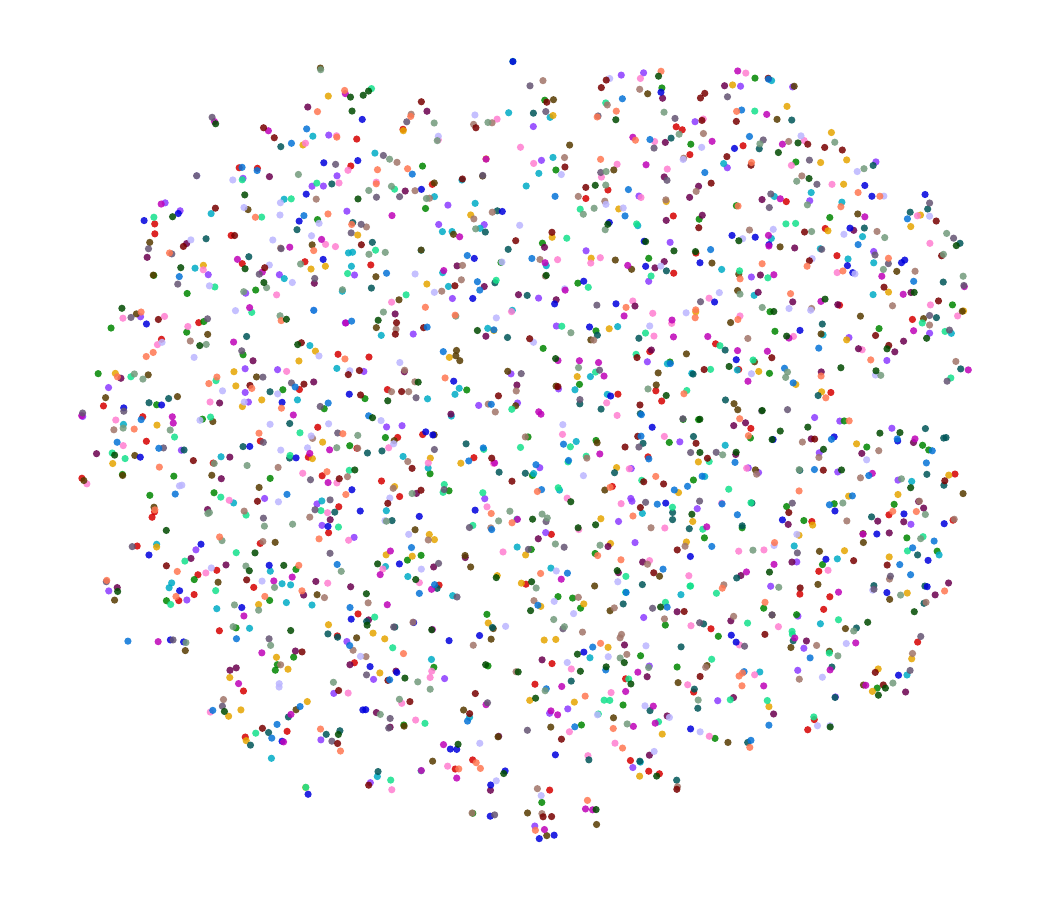}
        \\[2pt] (c)
    \end{minipage}

    \caption{\rev{t-SNE visualization of message representations learned on the MNIST dataset. Colors indicate samples from the same batch. (a) VQEL with $L=4$, (b) VQEL with $L=10$, and (c) VQEL + TTA with $L=10$. While increasing $L$ in VQEL yields a more dispersed representation space, substantial proximity between samples remains. TTA further enlarges inter-sample separation, resulting in a less ambiguous representation space and improved target discriminability.}}
    \label{fig:learned-language-space-mnist}
\end{figure*}

\begin{figure*}
    \centering

    \begin{minipage}{0.33\textwidth}
        \centering
        \includegraphics[width=\linewidth]{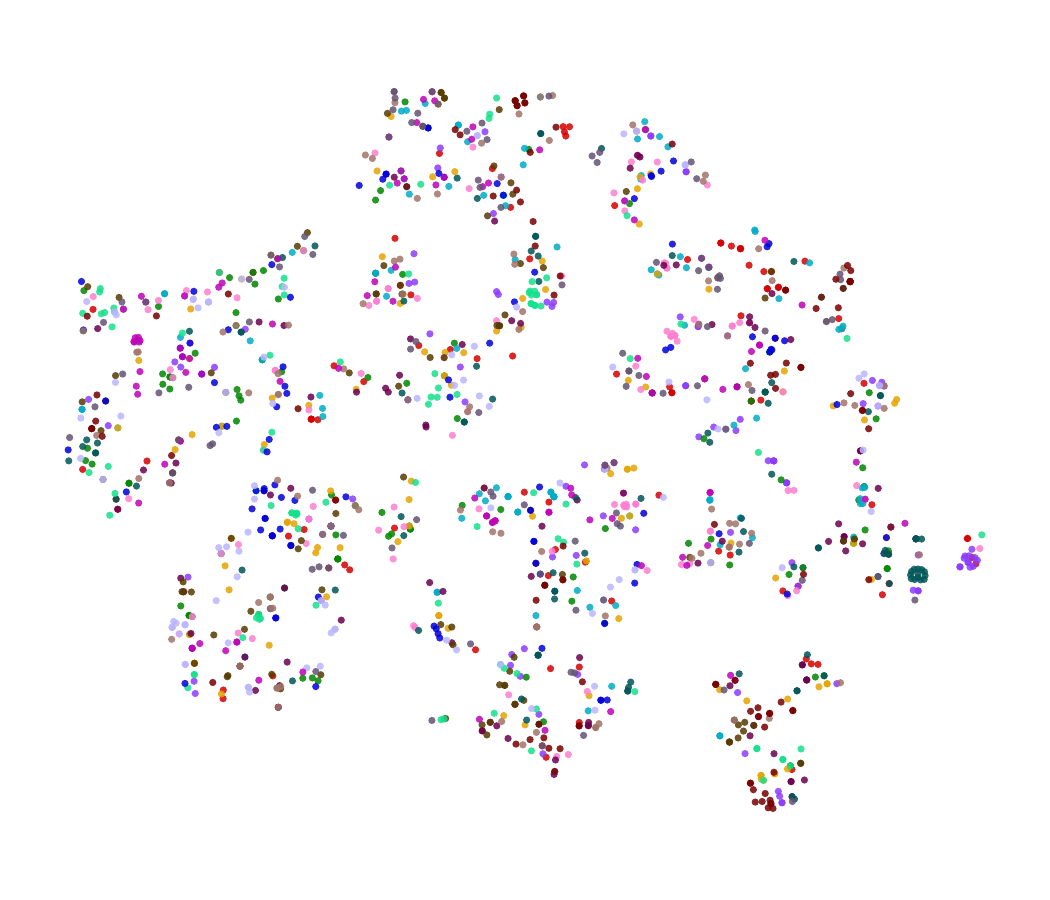}
        \\[2pt] (a)
    \end{minipage}
    \hfill
    \begin{minipage}{0.33\textwidth}
        \centering
        \includegraphics[width=\linewidth]{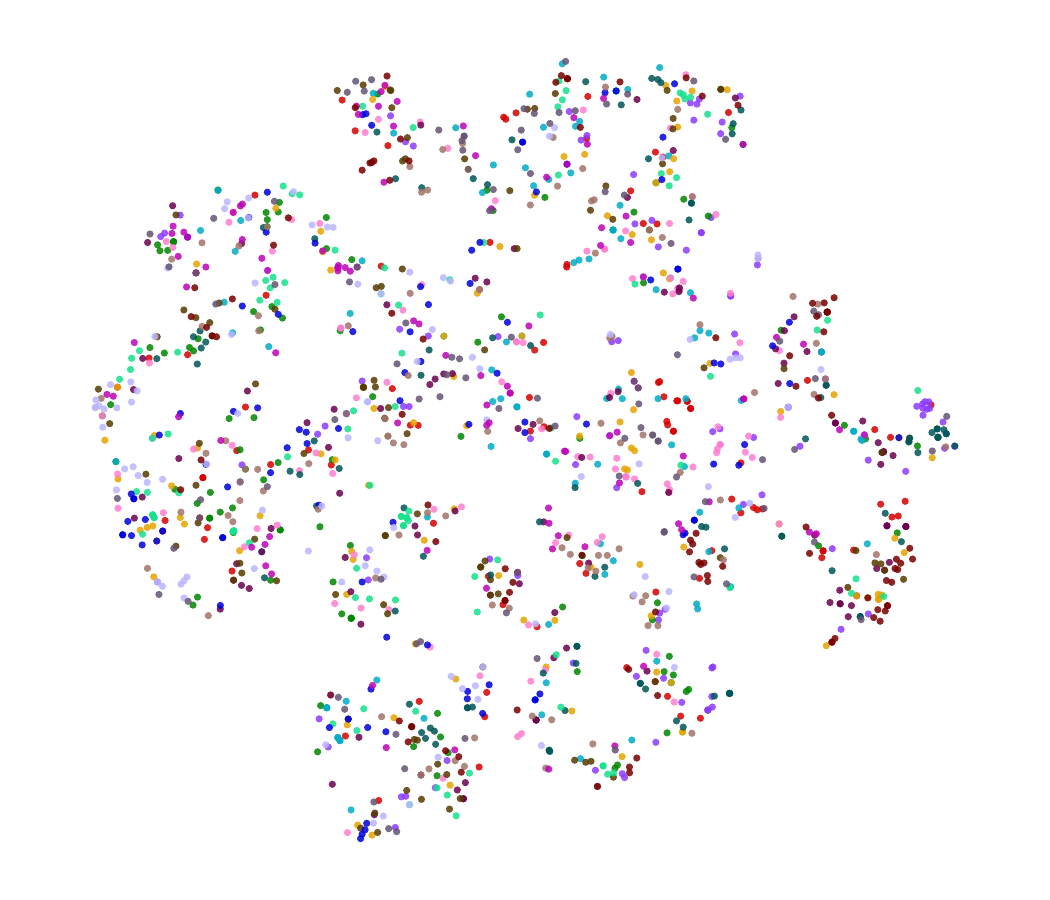}
        \\[2pt] (b)
    \end{minipage}
    \hfill
    \begin{minipage}{0.33\textwidth}
        \centering
        \includegraphics[width=\linewidth]{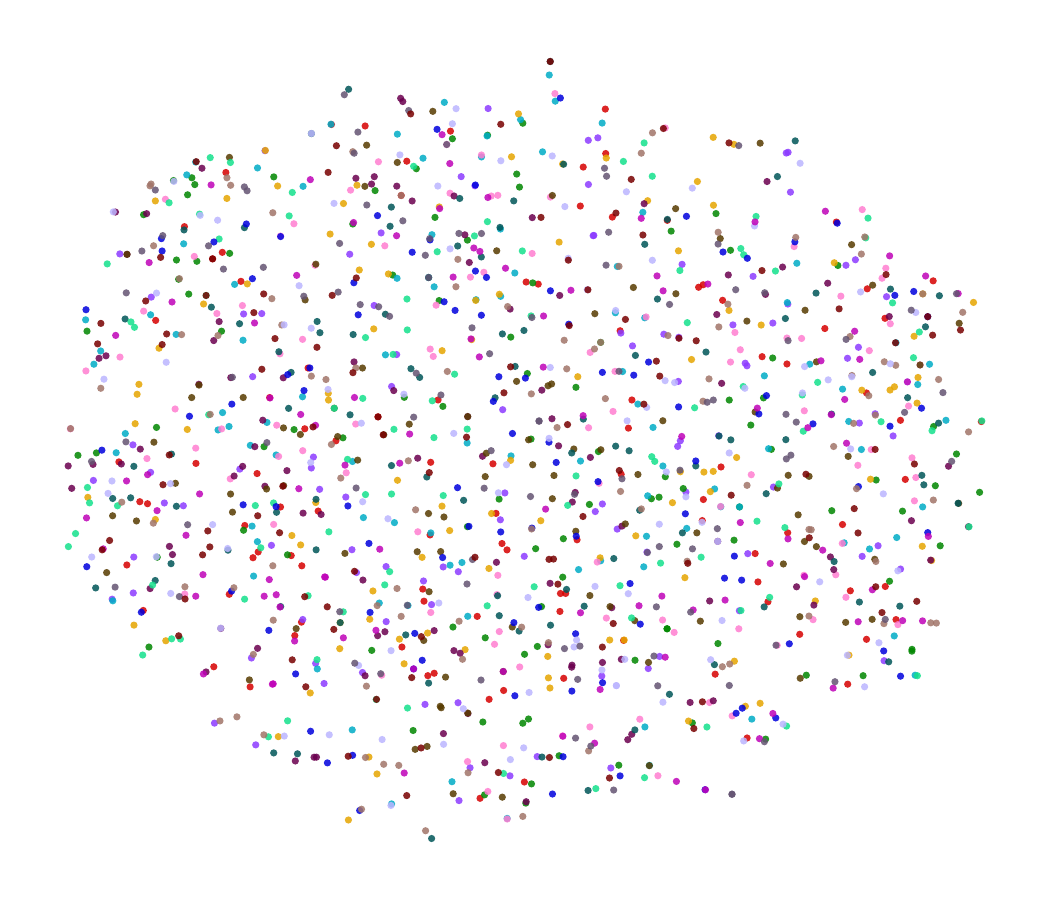}
        \\[2pt] (c)
    \end{minipage}

    \caption{\rev{t-SNE visualization of message representations learned on the ImageNet dataset. Colors denote samples belonging to the same batch. (a) VQEL with $L=4$, (b) VQEL with $L=5$, and (c) VQEL + TTA (dataset-level) with $L=5$. TTA encourages a more dispersed representation space, associated with better-separated target samples.}}
    \label{fig:learned-language-space-imagenet}
\end{figure*}

\rev{
To investigate the impact of TTA on the learned communication protocol, we analyze the emergent language representations before and after adaptation. Specifically, we visualize the message embeddings using t-SNE to compare the structure of the learned representation space.

Figures~\ref{fig:learned-language-space-mnist} and~\ref{fig:learned-language-space-imagenet} illustrate the t-SNE projections of the learned messages on MNIST and ImageNet, respectively. Before TTA, the learned representations exhibit substantial overlap, with many samples occupying nearby regions in the embedding space, making target discrimination more challenging. Following TTA, the message representations become more dispersed and better separated, consistent with the improved referential-game performance reported above. Additional qualitative examples of the generated messages before and after adaptation are provided in Appendix~\ref{app:qualitative-analysis}.
}

\section{Analysis}
In this section, we analyze the impact of the number of distractors, test-time learning rate, and the number of adaptation steps. For clarity, we report results on one representative dataset from each complexity category: MNIST for the multi-object complexity setting and ImageNet for the object-descriptive complexity setting.

\newrev{For each method, we use the setting-level message length selected by its respective criterion and reported in Table~\ref{tab:max-accuracies}.}

\subsection{Number of Distractors}
\label{sec:analysis-number-of-distractors}

\begin{figure*}
    \centering
    \begin{minipage}{0.48\textwidth}
        \centering
        \includegraphics[width=\linewidth]{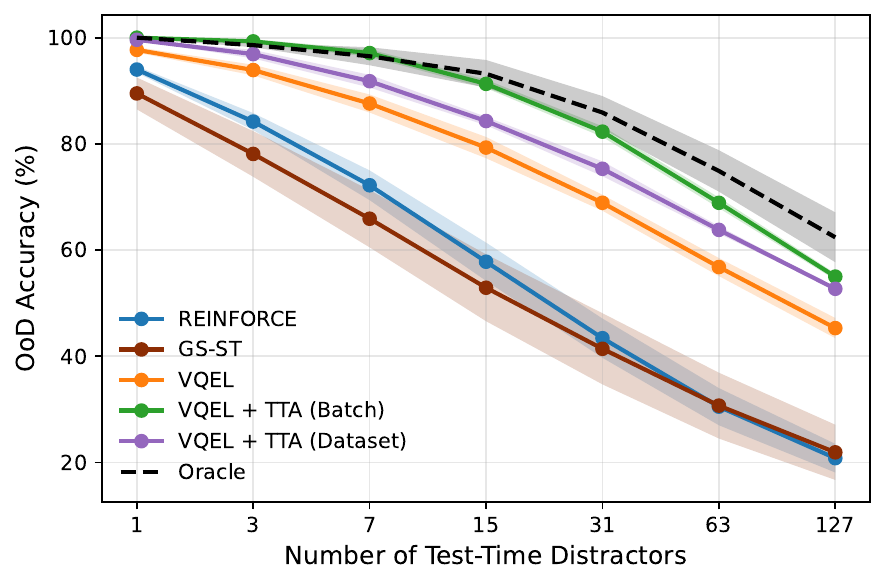}
    \end{minipage}
    \hfill
    \begin{minipage}{0.48\textwidth}
        \centering
        \includegraphics[width=\linewidth]{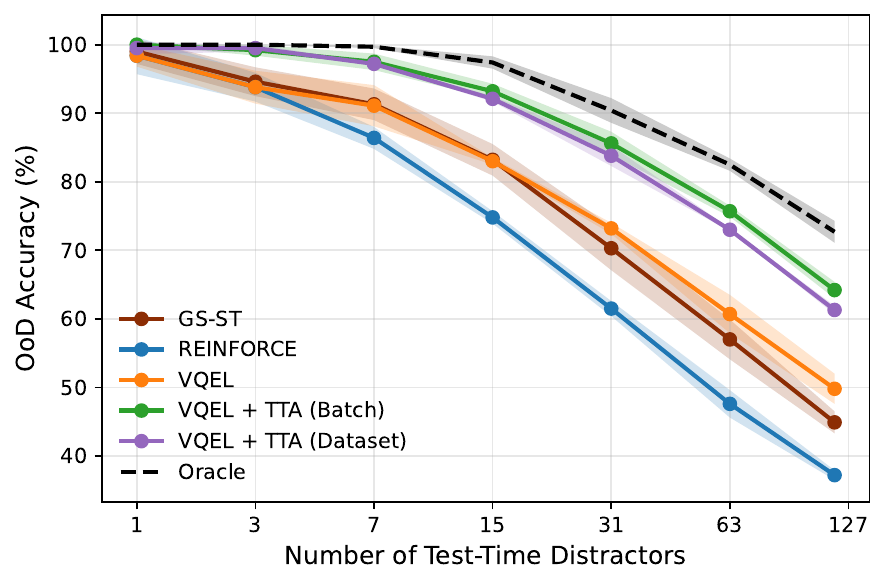}
    \end{minipage}
    \caption{\rev{TTA accuracy on the OoD split as a function of the number of distractors at test time. Left: MNIST dataset; Right: ImageNet dataset.}}
    \label{fig:number-of-distractors}
\end{figure*}

Figure \ref{fig:number-of-distractors} presents the accuracy of all five methods across varying numbers of distractors at test time. \newrev{Due to the limited number of images per class in our ImageNet subset (50 samples per class), we construct the complex OoD split using a different procedure from the one described in previous sections. Specifically, we group images with similar embeddings into the same batch, regardless of their class labels. Therefore, the ImageNet results in this experiment are not directly comparable to the main results.} Additional details on the construction of this split are provided in Appendix \ref{app:imagenet-ood-splits}.

As shown in the figure, TTA consistently improves performance across different numbers of distractors compared to the baseline methods. Moreover, TTA exhibits a smaller performance degradation as the number of distractors increases, indicating greater robustness in more challenging settings.

\subsection{Sensitivity to Test-Time Learning Rate and Adaptation Steps}

\begin{figure*}
    \centering
    \begin{minipage}{0.48\textwidth}
        \centering
        \includegraphics[width=\linewidth]{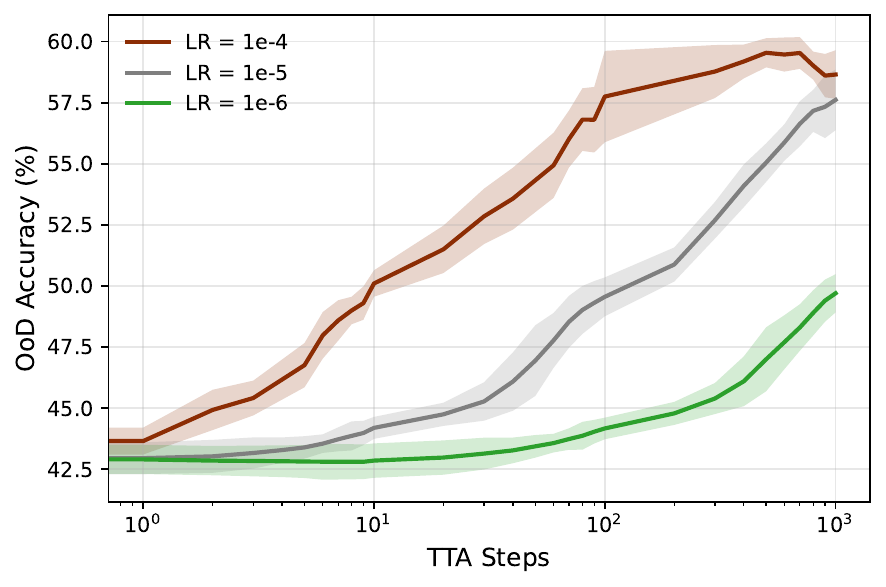}
    \end{minipage}
    \hfill
    \begin{minipage}{0.48\textwidth}
        \centering
        \includegraphics[width=\linewidth]{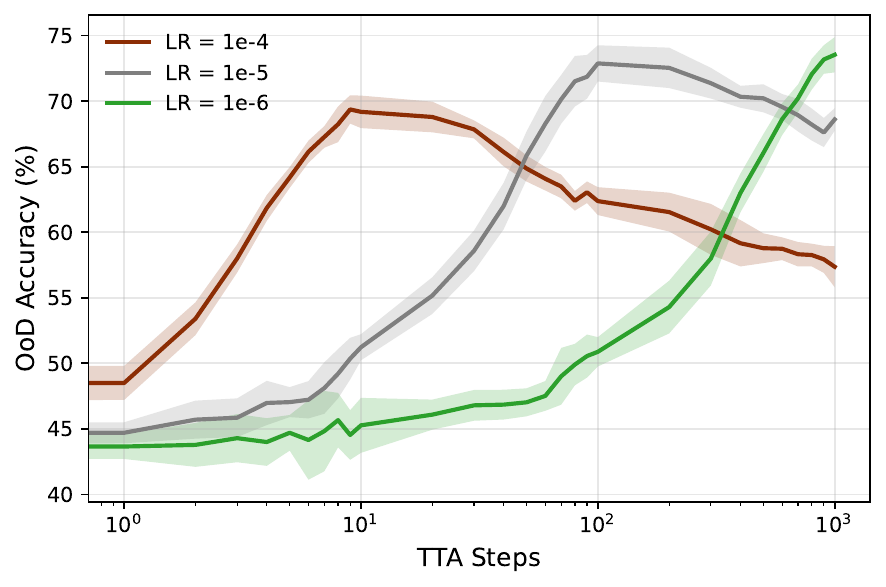}
    \end{minipage}
    \caption{Batch TTA accuracy on the OoD split across different test-time learning rates and numbers of adaptation steps. Left: MNIST dataset; Right: ImageNet dataset.}
    \label{fig:steps-batch-adaptation}
\end{figure*}

\begin{figure*}
    \centering
    \begin{minipage}{0.48\textwidth}
        \centering
        \includegraphics[width=\linewidth]{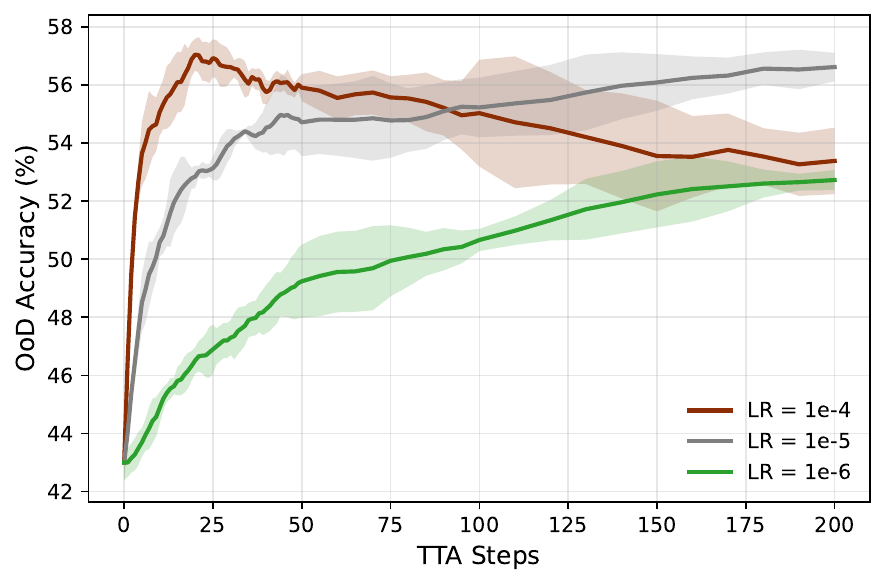}
    \end{minipage}
    \hfill
    \begin{minipage}{0.48\textwidth}
        \centering
        \includegraphics[width=\linewidth]{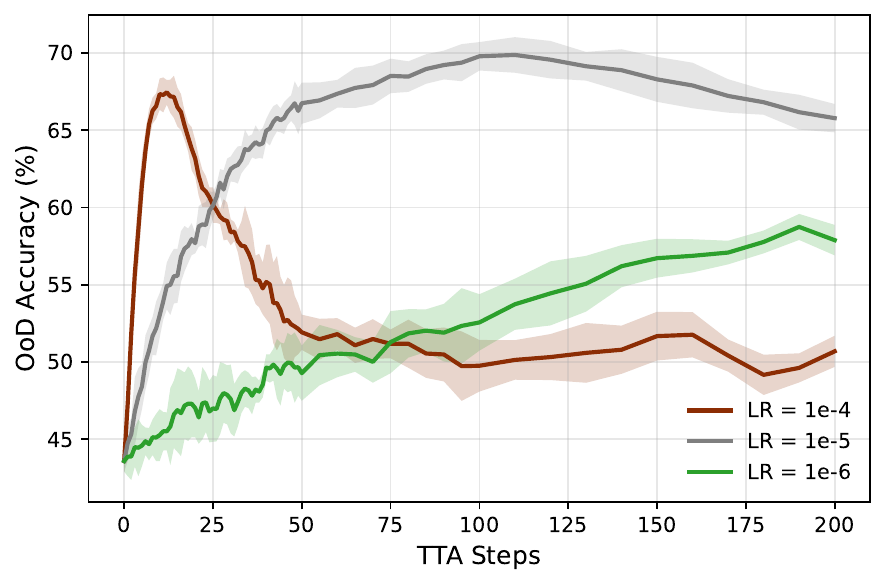}
    \end{minipage}
    \rev{\caption{Dataset TTA accuracy on the OoD split across different test-time learning rates and numbers of adaptation steps. Left: MNIST dataset; Right: ImageNet dataset.}}
    \label{fig:steps-dataset-adaptation}
\end{figure*}

Figures \ref{fig:steps-batch-adaptation} and \ref{fig:steps-dataset-adaptation} analyze the effect of the test-time learning rate and the number of adaptation steps on Batch TTA and Dataset TTA. As expected, smaller learning rates generally require a larger number of adaptation steps to achieve strong performance. We also observe that increasing the number of adaptation steps improves accuracy up to a certain point, after which performance begins to decline.
\rev{Note that while the optimal test-time learning rate is selected from the primary search space of $\{10^{-4}, 10^{-5}\}$, we additionally include $\eta = 10^{-6}$ in our analysis (Figures \ref{fig:steps-batch-adaptation} and \ref{fig:steps-dataset-adaptation}) as an ablation to demonstrate the performance degradation when the adaptation step size is overly conservative.}

This degradation occurs because the verifier serves only as an approximation of the receiver agent. Consequently, excessive optimization with respect to the verifier objective can lead to overfitting, which ultimately harms the model's performance.

\section{\rev{Discussion and Limitations}}
\label{sec:limitations}

\rev{While our proposed Test-Time Adaptation framework successfully mitigates communication failures under Complexity-OoD shifts, it introduces certain limitations and trade-offs that warrant transparent discussion.}

\subsection{\rev{Verifier-Receiver Alignment and Proxy Risks}}
\newrev{The core premise of our framework is that the sender's internal Message Perception Module can serve as a useful proxy for the external receiver's interpretation. To assess the relationship between these modules, we measured Centered Kernel Alignment (CKA) between the representations extracted from their perception modules after mutual play. CKA compares the geometric structure of learned representations and is invariant to orthogonal transformations of the embedding space, including feature permutations (see Appendix~\ref{app:cka_justification}). As shown in Table~\ref{tab:cka-alignment}, the CKA scores are high across all datasets ($>0.92$, $p < 0.001$) relative to a permutation-based random baseline (see Appendix~\ref{app:cka_baseline}). These results provide evidence of representational similarity, but they do not by themselves establish identical categorization boundaries or functional agreement under the OoD conditions used for adaptation. The remaining gap to the receiver-loss oracles further indicates that the internal verifier is informative but imperfect.}

\begin{table}[htbp]
\centering
\small
\setlength{\tabcolsep}{6pt}
\renewcommand{\arraystretch}{1.2}
\begin{tabular}{l c c c}
\toprule
\textbf{Dataset} & \textbf{CKA} & \textbf{Random Baseline} & \textbf{P-value} \\
\midrule
MNIST           & $0.965 \pm 0.004$ & $0.004 \pm 0.001$  & $< 0.001$ \\
ShapeWorld      & $0.946 \pm 0.015$ & $0.003 \pm 0.001$  & $< 0.001$ \\
ImageNet        & $0.950 \pm 0.014$ & $0.006 \pm 0.001$  & $< 0.001$ \\
COCO            & $0.924 \pm 0.005$ & $0.005 \pm 0.001$  & $< 0.001$ \\
\bottomrule
\end{tabular}
\caption{\rev{Centered Kernel Alignment (CKA) similarity between the sender’s and receiver’s message representations.}}
\label{tab:cka-alignment}
\end{table}

\rev{Despite this strong global correlation, relying on an internal verifier introduces several deployment risks. First, under severe distribution shifts, the verifier may diverge from the receiver's actual interpretation. In such cases, adapting the sender could lead to self-overfitting, where the sender generates idiosyncratic or overly verbose messages that confidently satisfy its internal verifier but remain ambiguous to the external receiver, ultimately causing a \textit{receiver-side mismatch}. Exploring lightweight adaptation on the receiver side to mitigate this mismatch represents a critical avenue for future work.
To characterize the performance headroom of our approach, we compare it against an oracle that adapts using the receiver's loss under a matched computational budget (Section~\ref{sec:baselines}), as well as a stronger oracle that updates all sender and receiver components (Appendix~\ref{app:oracles}). These oracles bound the achievable gains and quantify the gap left by relying solely on the sender's internal verifier.
\newrev{The remaining gap to the receiver-loss oracle, and especially to the stronger oracle that adapts all sender and receiver components, is a practical upper-bound reference for the current sender-only formulation. It quantifies the performance headroom associated with replacing direct receiver feedback by the internal verifier and restricting adaptation to the sender's message generator. These oracle variants are not directly deployable in our one-shot setting because they rely on receiver-side loss signals or updates; nevertheless, the gap shows that the internal verifier is an informative but imperfect proxy. Closing it may require better-calibrated verifiers, lightweight receiver-side adaptation, or joint adaptation under additional deployment assumptions.}
Furthermore, it is important to distinguish the risk of \textit{catastrophic forgetting} between our two adaptation settings. While our batch-level TTA naturally avoids this issue by resetting the generator's weights to the base model for each test instance, the continuous optimization involved in dataset-level adaptation poses a potential risk. In a continual deployment stream, adapting solely on OoD instances might cause the sender to lose its established communication protocol for simple ID scenarios unless careful regularization is applied.}
\newrev{In the current sender-only formulation, the receiver remains fixed to preserve one-shot communication and isolate the effect of sender-side adaptation. However, a receiver trained mainly on shorter training-time messages may not fully accommodate substantially longer adapted messages under severe shifts. Lightweight receiver-side adaptation, while preserving the one-shot communication constraint, is therefore an important direction for future work.}
\newrev{More generally, the fixed baselines and TTA do not use identical test-time budgets. Batch-level TTA can use the actual candidate set, whereas dataset-level TTA uses an unlabeled test stream; both variants also perform additional optimization at deployment. We therefore interpret the reported gains as the benefit of exploiting these available test-time signals through a learned verifier under an expanded information and computation budget, rather than as a strictly apples-to-apples feed-forward comparison.}
\newrev{We also do not evaluate substantially more severe or heterogeneous distribution shifts in which the sender verifier may diverge more strongly from the receiver. The current results should therefore not be interpreted as evidence that the verifier remains reliable under arbitrary deployment conditions. Evaluating such shifts and directly measuring the resulting verifier--receiver mismatch are important directions for future work.}

\subsection{\rev{Computational Overhead}}
\newrev{A fundamental limitation of performing active optimization at deployment is the increased computational cost. Tables \ref{tab:cost-mnist} and \ref{tab:cost-imagenet} report inference latency, memory usage, and GFLOPs on the MNIST and ImageNet datasets, respectively. These measurements were obtained on an NVIDIA GeForce RTX 4090 GPU. Unlike standard feed-forward inference seen in the baselines, our TTA framework requires multiple forward and backward passes (e.g., up to hundreds of gradient steps) per test instance. Consequently, while TTA yields critical accuracy gains in ambiguous scenarios, the resulting accumulated computational overhead makes it impractical for strict real-time systems or resource-constrained edge applications. The early-stopping, single-step, and forward-only alternatives discussed in the method section are not evaluated here; their ability to reduce this overhead without sacrificing accuracy remains an open empirical question and an important direction for future work.}

\begin{table}[htbp]
\centering
\small
\setlength{\tabcolsep}{3pt}
\renewcommand{\arraystretch}{1.2}
\newrev{
\begin{tabular}{l c c c c}
\toprule
\textbf{Method} & \textbf{Accuracy (\%)} & \textbf{Latency (ms)} & \textbf{Mem (GB)} & \textbf{GFLOPs} \\
\midrule
GS-ST                     & $16.6_{\pm 4.7}$  & $0.13_{\pm 0.01}$ & $0.09$ & $4.1$ \\
REINFORCE                 & $14.2_{\pm 7.2}$  & $0.12_{\pm 0.00}$ & $0.09$ & $4.1$ \\
VQEL                      & $43.0_{\pm 0.6}$  & $0.15_{\pm 0.00}$ & $0.09$ & $4.1$ \\
TTA (Dataset)             & $55.5_{\pm 0.7}$  & $4.89_{\pm 0.14}$ & $0.34$ & $5.5$ \\
TTA (Batch)               & $57.4_{\pm 1.1}$  & $43.82_{\pm 3.73}$& $0.64$ & $17.6$ \\
\bottomrule
\end{tabular}
}
\caption{\newrev{Computational cost on the \textbf{MNIST} dataset for $L=10$ setting. Latency is reported per instance. Mem and GFLOPs denote the total memory footprint and cumulative floating-point operations required to process the entire OoD test split (2,000 instances).}}
\label{tab:cost-mnist}
\end{table}

\begin{table}[htbp]
\centering
\small
\setlength{\tabcolsep}{3pt}
\renewcommand{\arraystretch}{1.2}
\newrev{
\begin{tabular}{l c c c c}
\toprule
\textbf{Method} & \textbf{Accuracy (\%)} & \textbf{Latency (ms)} & \textbf{Mem (GB)} & \textbf{GFLOPs} \\
\midrule
GS-ST                     & $37.6_{\pm 4.7}$ & $0.20_{\pm 0.04}$  & $0.56$ & $0.8$  \\
REINFORCE                 & $33.7_{\pm 1.2}$ & $0.21_{\pm 0.08}$  & $0.56$ & $0.8$  \\
VQEL                      & $43.6_{\pm 0.7}$ & $0.19_{\pm 0.00}$  & $0.57$ & $0.8$  \\
TTA (Dataset)             & $69.4_{\pm 0.3}$ & $7.59_{\pm 0.09}$ & $1.20$ & $25.2$ \\
TTA (Batch)               & $68.9_{\pm 0.8}$ & $7.57_{\pm 0.04}$ & $1.82$ & $25.2$ \\
\bottomrule
\end{tabular}
}
\caption{\newrev{Computational cost on the \textbf{ImageNet} dataset for $L=5$ setting. Latency is reported per instance. Mem and GFLOPs denote the total memory footprint and cumulative floating-point operations required to process the entire OoD test split (1,600 instances).}}
\label{tab:cost-imagenet}
\end{table}

\subsection{\rev{Dependence on Self-Play Architectures}}
\rev{Our test-time adaptation mechanism is intrinsically tied to architectures that foster an internal verifier. Specifically, it builds upon the VQEL framework, as its self-play phase naturally yields a grounded Message Perception Module. Consequently, our approach cannot be directly or trivially applied to standard emergent communication baselines (e.g., vanilla REINFORCE or GS-ST) without significant structural modifications to inject self-supervised alignment capabilities.}
\newrev{Accordingly, the present results should be understood as a VQEL-based instantiation of a broader sender-side TTA principle, rather than as evidence that the current implementation is a drop-in method for all emergent-communication architectures. Extending the approach to other architectures would require constructing or learning a suitable self-supervised verifier, together with an objective that connects the verifier to their discrete message generators. We leave this architectural extension and its experimental validation for future work.}

\section{Conclusion}
\newrev{When test episodes become inherently more difficult than training scenarios—due to the presence of multiple objects or highly confusable distractors—standard emergent communication protocols often fail, suffering from underspecified and ambiguous messages. In this paper, we study this broader family of deployment-time shifts, with the multi-object settings providing a direct Complexity-OoD evaluation and the ImageNet and COCO settings representing object-descriptive or candidate-confusability shifts. We introduce a sender-side Test-Time Adaptation (TTA) framework to address these challenges. By repurposing the sender's internal message perception module as an inference-time verifier, our approach optimizes the message generation process instance-by-instance without requiring receiver-side gradients or extra interactive dialogue. Empirical evaluations across MNIST, ShapeWorld, ImageNet, and COCO demonstrate that TTA consistently enhances communication robustness, particularly when longer descriptions are useful for disambiguating complex candidate sets.}


\appendices
\section{Datasets}
\label{app:datasets}
\begin{figure*}[h!]
    \centering
    \begin{minipage}{0.35\textwidth}
        \centering
        \includegraphics[width=\linewidth]{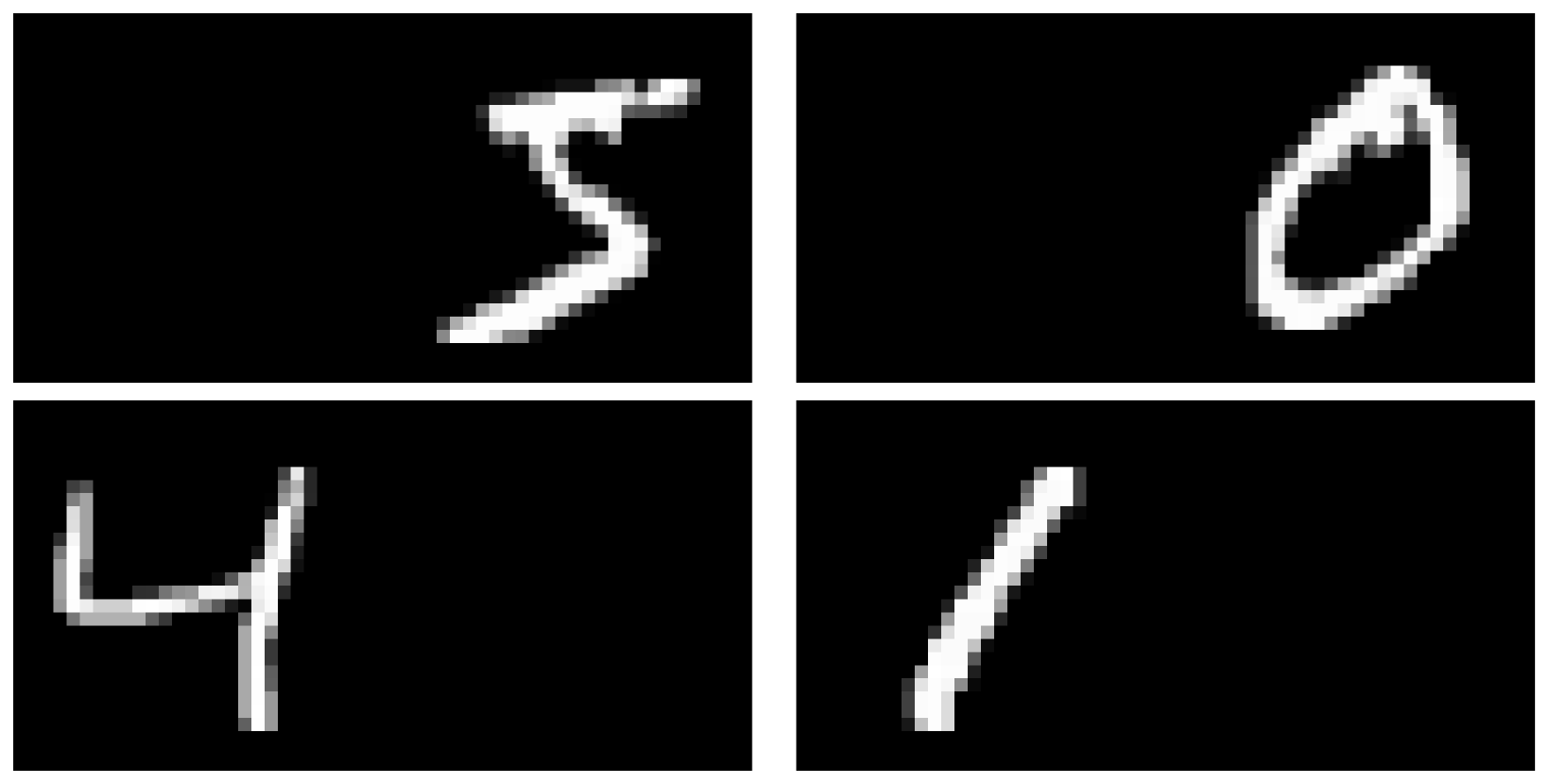}
        \\[2pt] (a)
    \end{minipage}
    \hspace{0.02\textwidth}
    \begin{minipage}{0.35\textwidth}
        \centering
        \includegraphics[width=\linewidth]{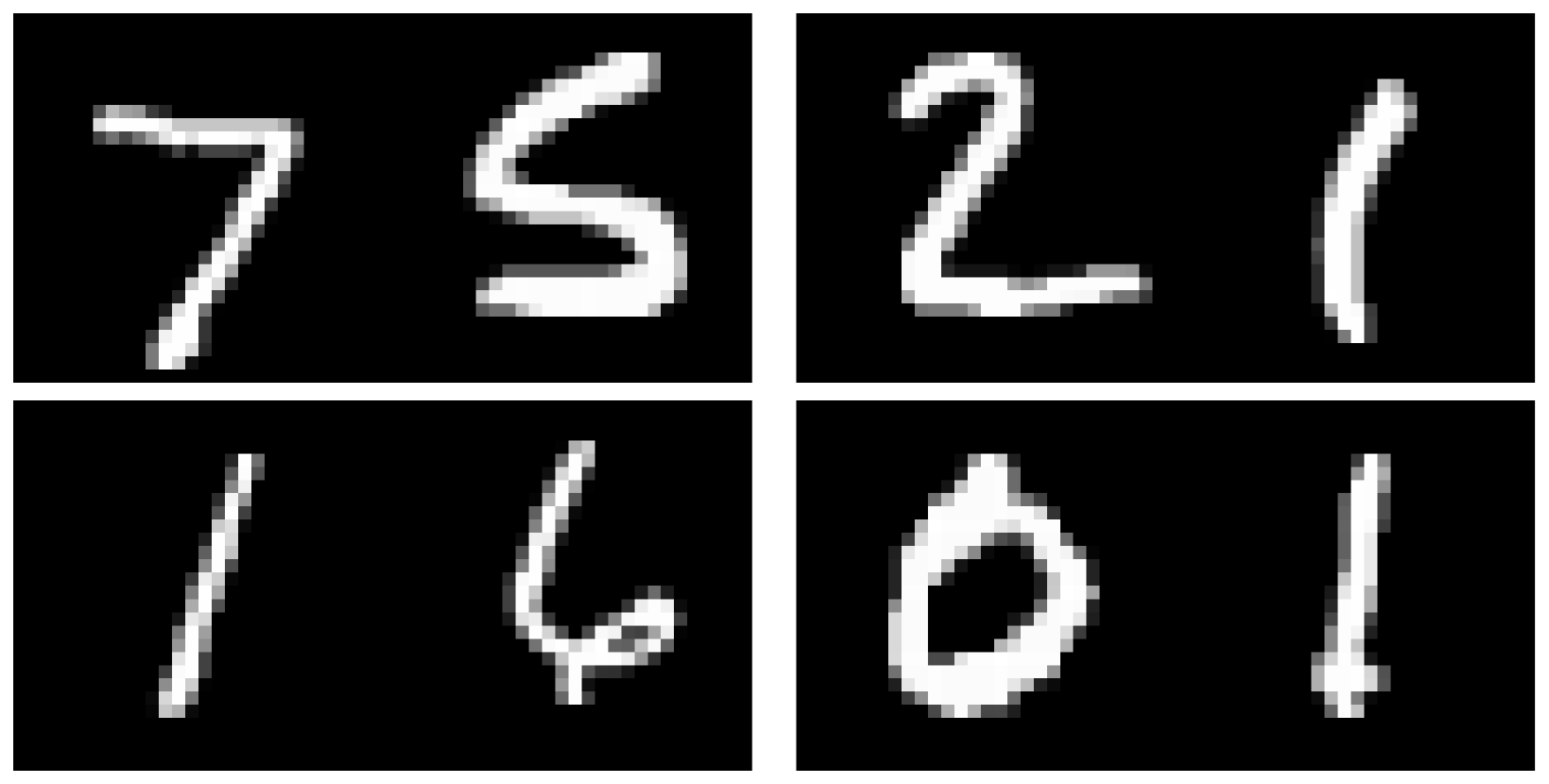}
        \\[2pt] (b)
    \end{minipage}
    \caption{Examples from the MNIST dataset. (a) 1-Digit subset (ID) 
    used for training. (b) 2-Digit subset (OoD) used for test-time evaluation.}
    \label{fig:mnist_samples}
\end{figure*}

\begin{figure*}[h!]
    \centering
    \begin{minipage}{0.35\textwidth}
        \centering
        \includegraphics[width=\linewidth]{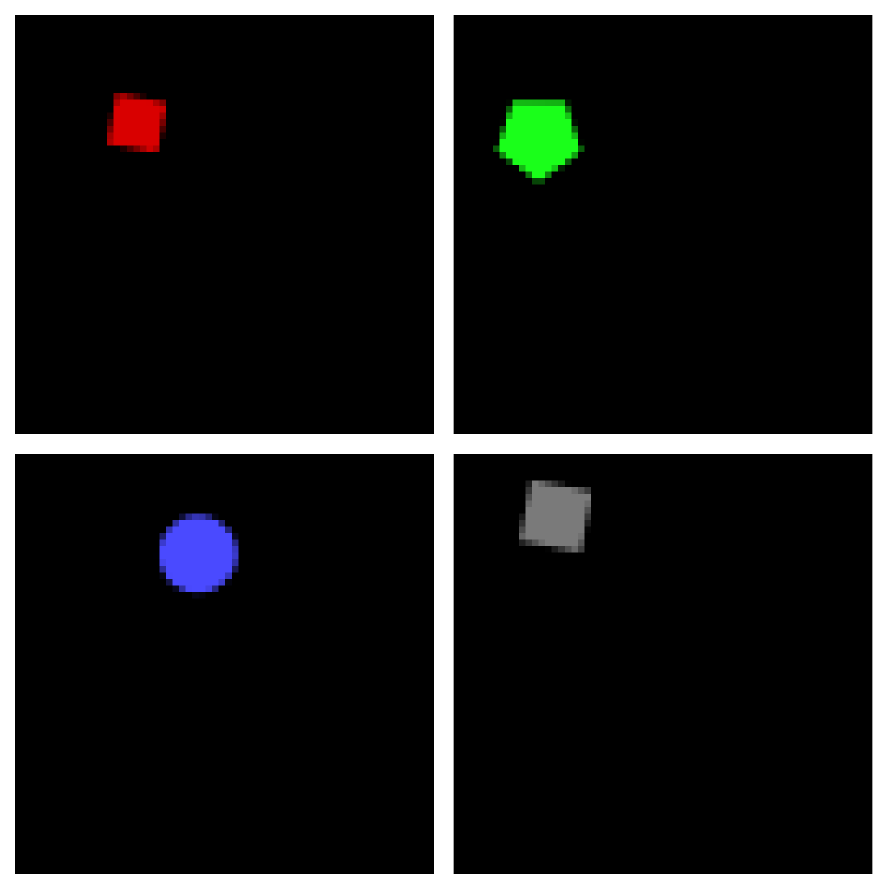}
        \\[2pt] (a)
    \end{minipage}
    \hspace{0.02\textwidth}
    \begin{minipage}{0.35\textwidth}
        \centering
        \includegraphics[width=\linewidth]{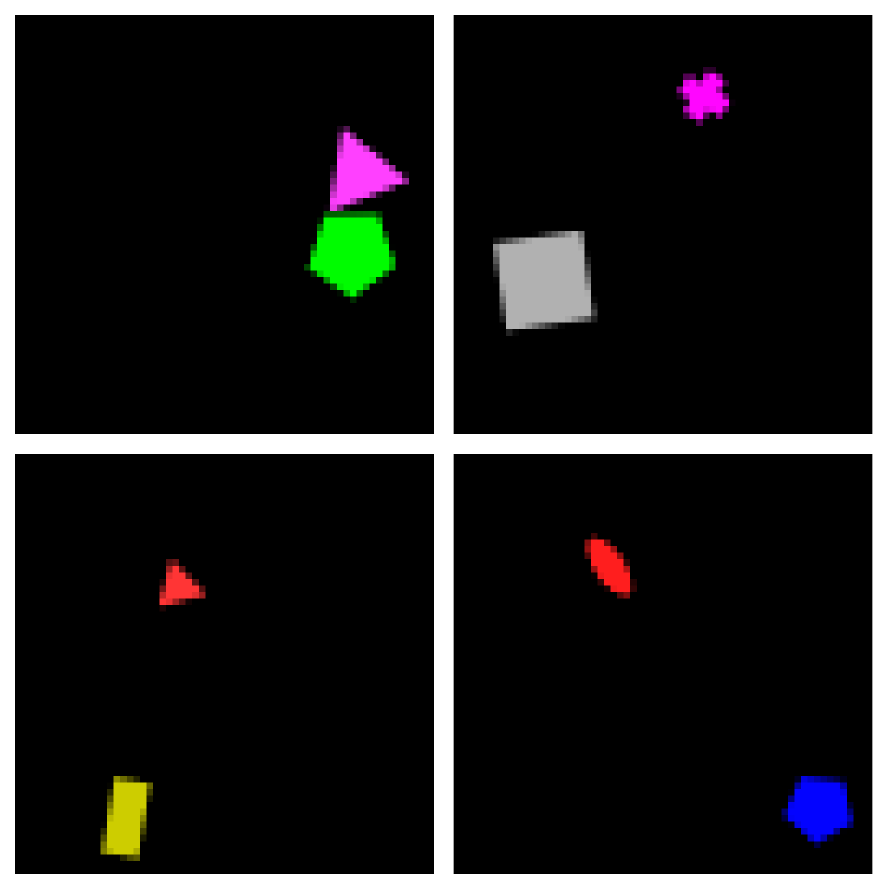}
        \\[2pt] (b)
    \end{minipage}
    \caption{Examples from the ShapeWorld dataset. (a) Single-object subset (ID) 
    used for training. (b) Two-object subset (OoD) used for test-time evaluation.}
    \label{fig:shape_samples}
\end{figure*}

\begin{figure*}[h!]
    \centering
    \begin{minipage}{0.35\textwidth}
        \centering
        \includegraphics[width=\linewidth]{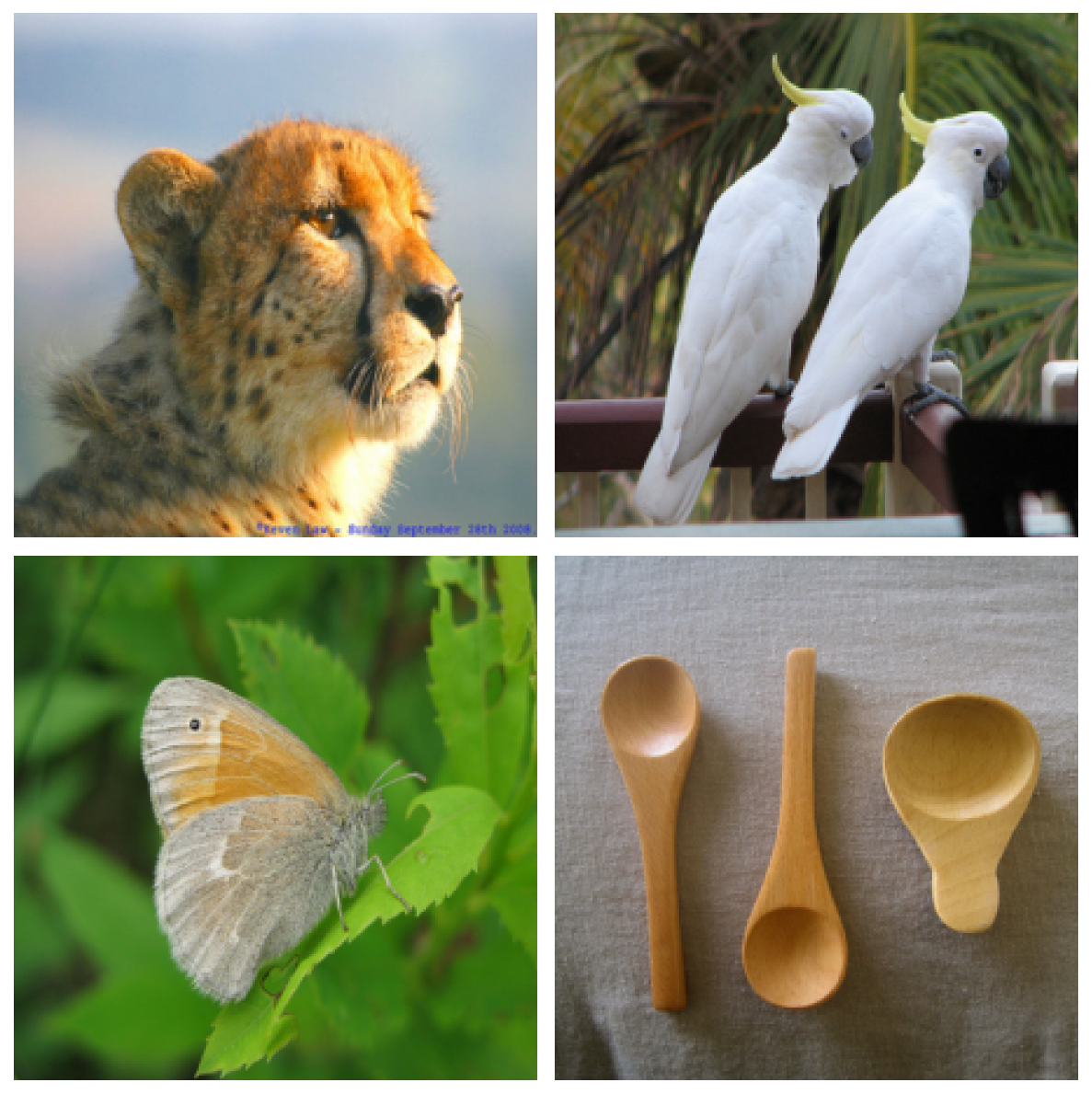}
        \\[2pt] (a)
    \end{minipage}
    \hspace{0.02\textwidth}
    \begin{minipage}{0.35\textwidth}
        \centering
        \includegraphics[width=\linewidth]{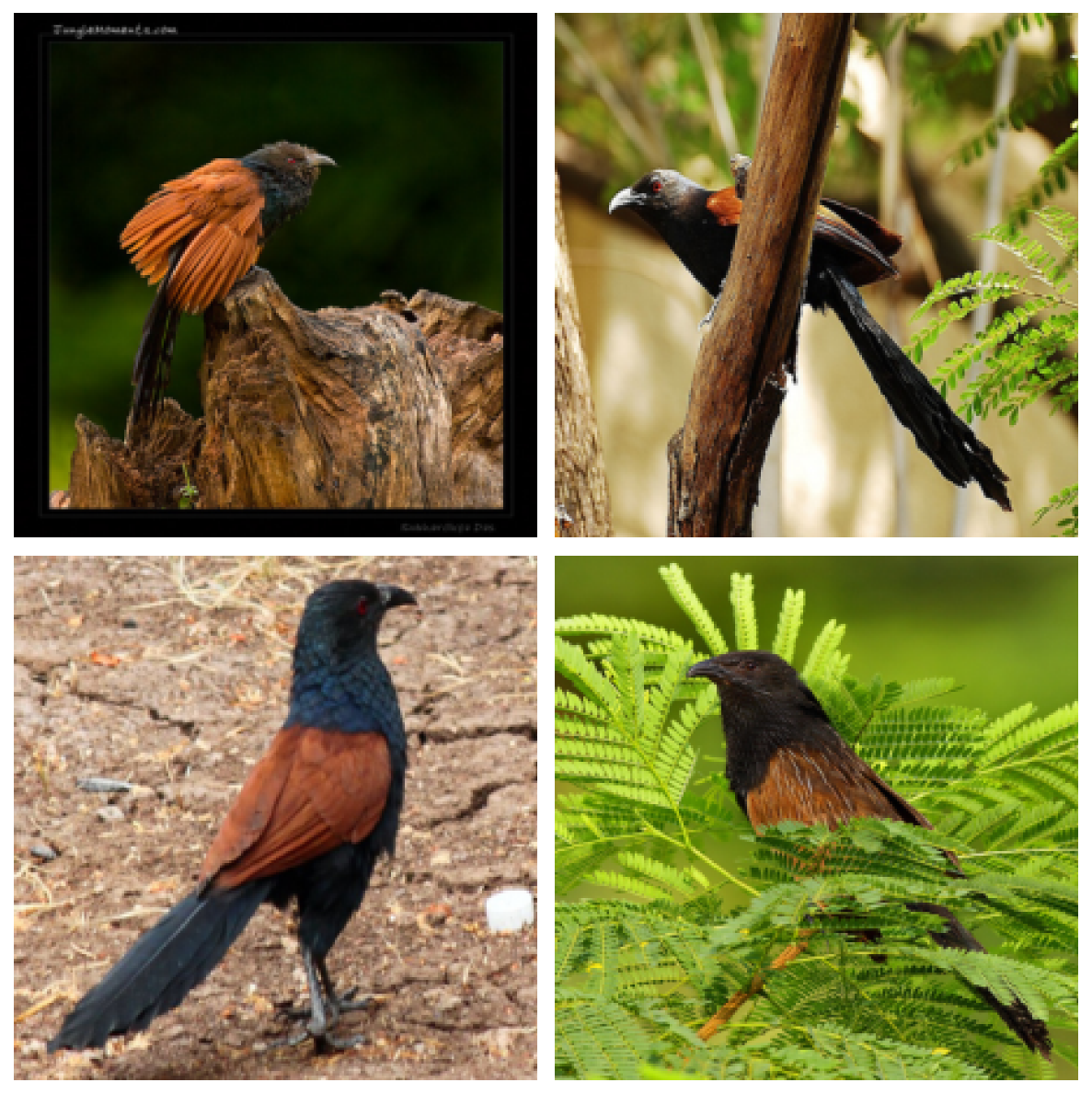}
        \\[2pt] (b)
    \end{minipage}
    \caption{Examples from the ImageNet dataset. (a) Standard subset (ID) used for 
    training. (b) Complex subset (OoD) used for test-time evaluation.}
    \label{fig:imagenet_samples}
\end{figure*}

\begin{figure*}[h!]
    \centering
    \begin{minipage}{0.35\textwidth}
        \centering
        \includegraphics[width=\linewidth]{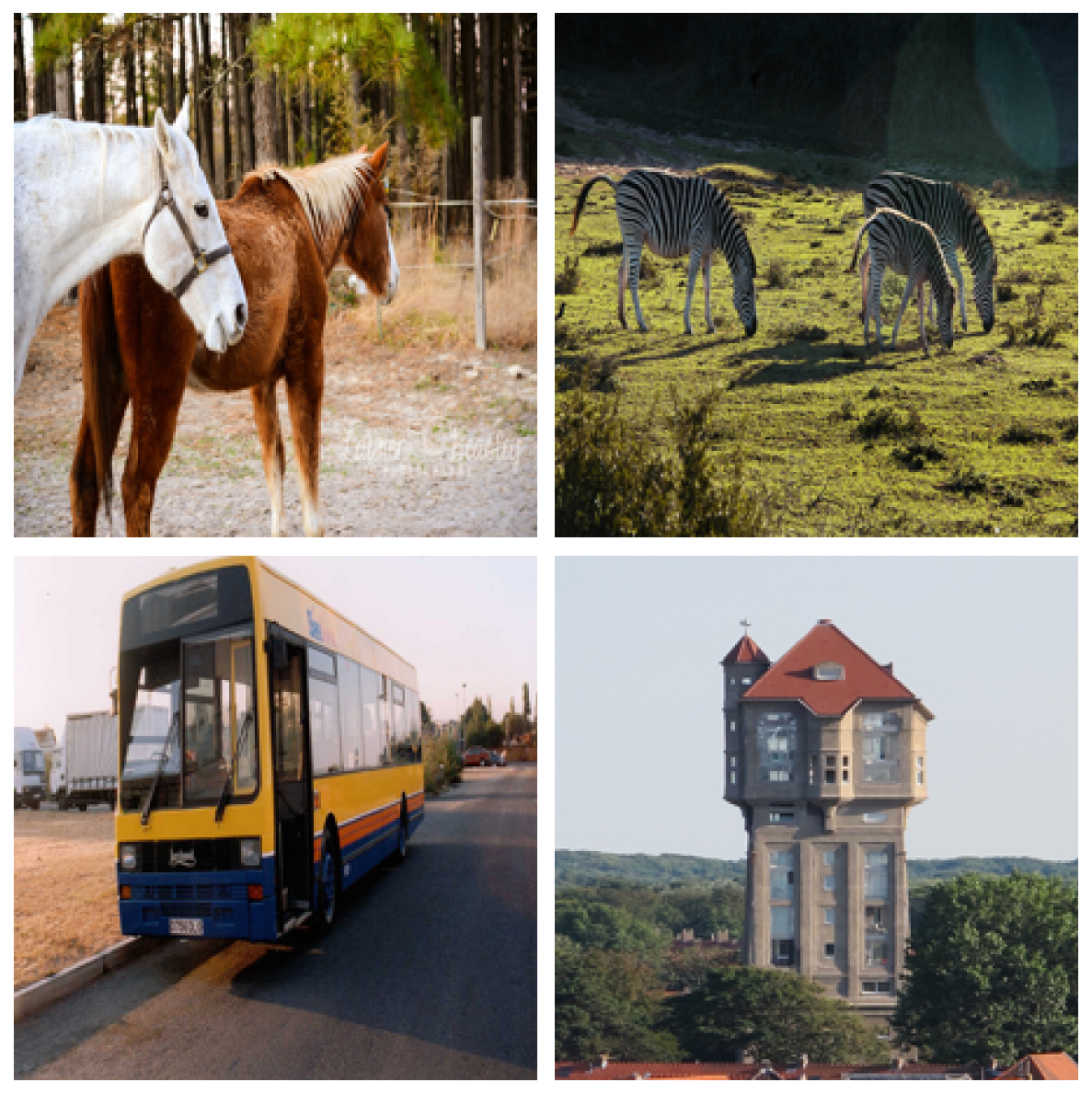}
        \\[2pt] (a)
    \end{minipage}
    \hspace{0.02\textwidth}
    \begin{minipage}{0.35\textwidth}
        \centering
        \includegraphics[width=\linewidth]{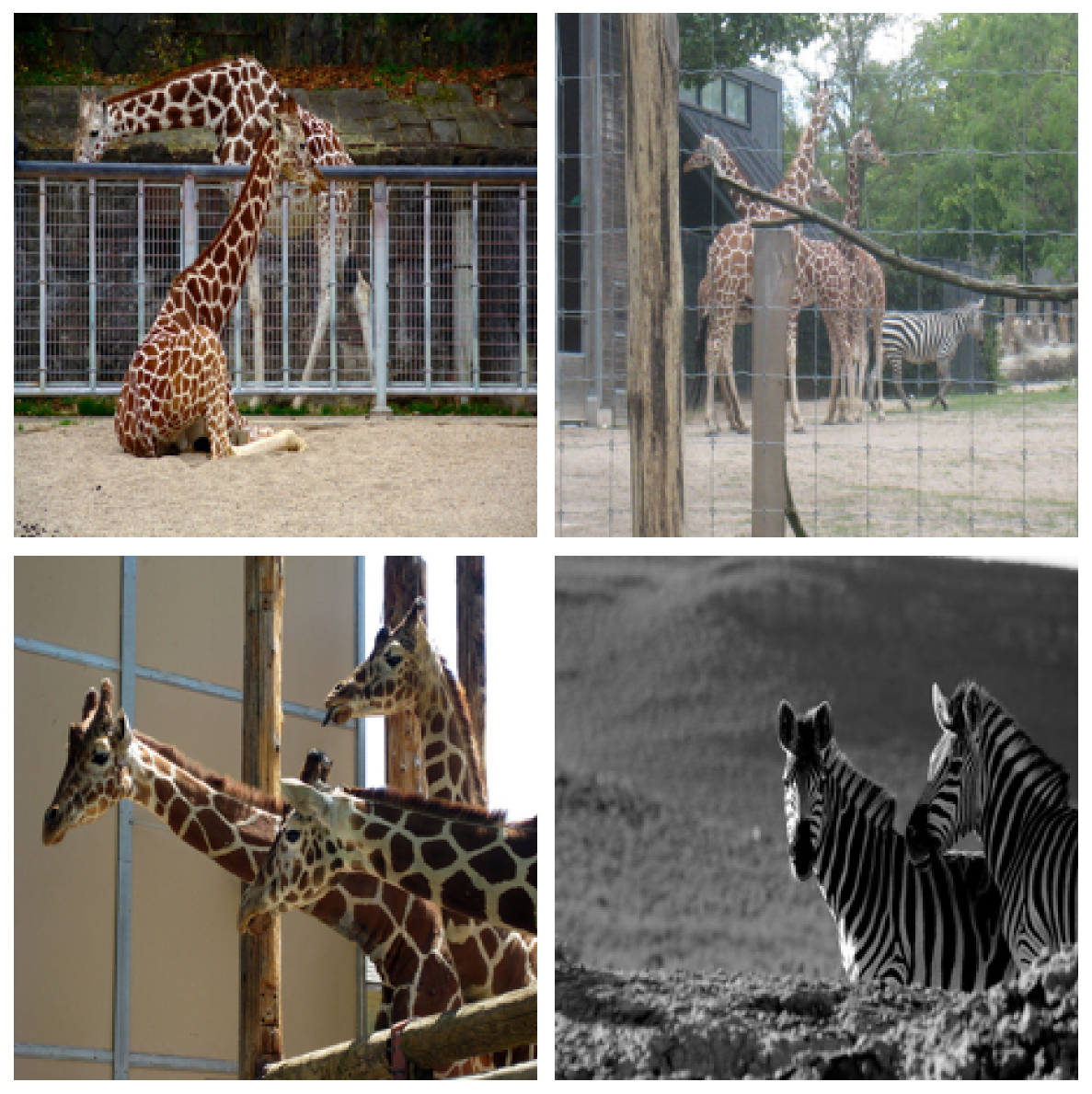}
        \\[2pt] (b)
    \end{minipage}
    \caption{\rev{Examples from the COCO dataset. (a) Standard subset (ID) used for training. (b) Complex subset (OoD) used for test-time evaluation. These examples exhibit similar embedding representations while potentially belonging to different semantic classes.}}
    \label{fig:coco_samples}
\end{figure*}

Some example images from the MNIST, ShapeWorld, ImageNet \rev{and COCO} datasets are shown in Figures~\ref{fig:mnist_samples}, \ref{fig:shape_samples}, \ref{fig:imagenet_samples}, \rev{and \ref{fig:coco_samples}}, respectively.

\section{\newrev{Visual Genome: Splits and Results}}
\label{app:genome}

\newrev{We construct the Visual Genome splits to increase scene complexity while avoiding overlap with COCO. We first exclude images originating from COCO and partition the remaining images according to the number of annotated regions. Images with fewer regions form the ID (Standard) split, while those with more regions form the OoD (Complex) split. The resulting splits contain, on average, 42.03 and 88.23 regions per image, respectively, with the OoD split having approximately twice the region density of the ID split. Evaluation results are reported in Table~\ref{tab:genome-ood}. VQEL exhibits a $10.1\%$ absolute accuracy drop from ID to OoD, demonstrating the increased difficulty of the Complex split.}

\begin{table}
\centering
\setlength{\tabcolsep}{7pt}
\renewcommand{\arraystretch}{1.1}
\newrev{
\begin{tabular}{l c c}
\toprule
\multirow{2}{*}{\textbf{Method}} 
& \multicolumn{2}{c}{\textbf{VisualGenome}} \\
\cmidrule(lr){2-3}
& Standard (ID) & Complex (OoD) \\
\midrule
GS-ST      & $61.6_{\pm 1.6}$ & $53.1_{\pm 0.8}$ \\
REINFORCE  & $55.3_{\pm 5.3}$ & $44.1_{\pm 6.3}$ \\
VQEL       & $63.4_{\pm 2.4}$ & $53.3_{\pm 2.6}$ \\
\bottomrule
\end{tabular}
}
\caption{\newrev{Test accuracy (\%) of the evaluated methods on the ID and OoD splits of the Visual Genome dataset.}}
\label{tab:genome-ood}
\end{table}

\newrev{We further evaluate TTA on the Complex split, with results shown in Table~\ref{tab:genome-best-lenght}. We use a learning rate of $10^{-5}$ and select the number of adaptation steps using an elbow criterion over $\{5,10,15,20,25\}$ for dataset-level adaptation and $\{25,50,100,150,200\}$ for batch-level adaptation. We similarly select one message length per TTA variant from 4--15 using verifier scores aggregated over the full OoD split, following the setting-level procedure used in the main experiments. TTA improves VQEL accuracy by $10.6$ and $11.7$ percentage points for dataset- and batch-level adaptation, respectively, largely recovering the OoD performance degradation.}

\begin{table}
\centering
\small
\setlength{\tabcolsep}{6pt}
\renewcommand{\arraystretch}{1.15}
\newrev{
\begin{tabular}{l c c}
\toprule
\textbf{Method} & \textbf{TT Msg. Len} & \textbf{Accuracy} \\
\midrule
GS-ST                        & $5$          & $53.9_{\pm 1.9}$ \\
REINFORCE                    & $5$          & $47.2_{\pm 7.9}$ \\
VQEL                         & $6$          & $57.2_{\pm 0.9}$ \\
\textbf{VQEL+TTA (Batch)}    & $\mathbf{7}$ & $\mathbf{65.0_{\pm 0.9}}$ \\
VQEL+TTA (Dataset)           & ${6}$        & ${63.9_{\pm 2.7}}$ \\
\bottomrule
\end{tabular}
}
\caption{\newrev{Setting-level selected message length and corresponding mean OoD accuracy on Visual Genome. One length is reported per method for the full OoD split. VQEL and the TTA variants use an elbow-based criterion on aggregate sender-side self-play or verifier scores; REINFORCE and GS-ST use mutual-play accuracy as an oracle reference for length selection. TTA achieves higher accuracy than all baselines.}}
\label{tab:genome-best-lenght}
\end{table}

\section{\rev{Ablation Studies}}
\subsection{\rev{Full Sender Adaptation vs. Generator-Only Adaptation}}
\label{app:full-adaptation}

\rev{A natural alternative to our proposed approach is \textit{full sender adaptation}, where all modules of the sender (Object Perception, Message Generator, and Message Perception) are simultaneously fine-tuned at test time using the internal self-play loss. To isolate the contribution of our specific design choice, we conducted an ablation study comparing our generator-only adaptation against full sender adaptation.}

\begin{figure*}[htbp]
    \centering
    \begin{minipage}{0.48\textwidth}
        \centering
        \includegraphics[width=\linewidth]{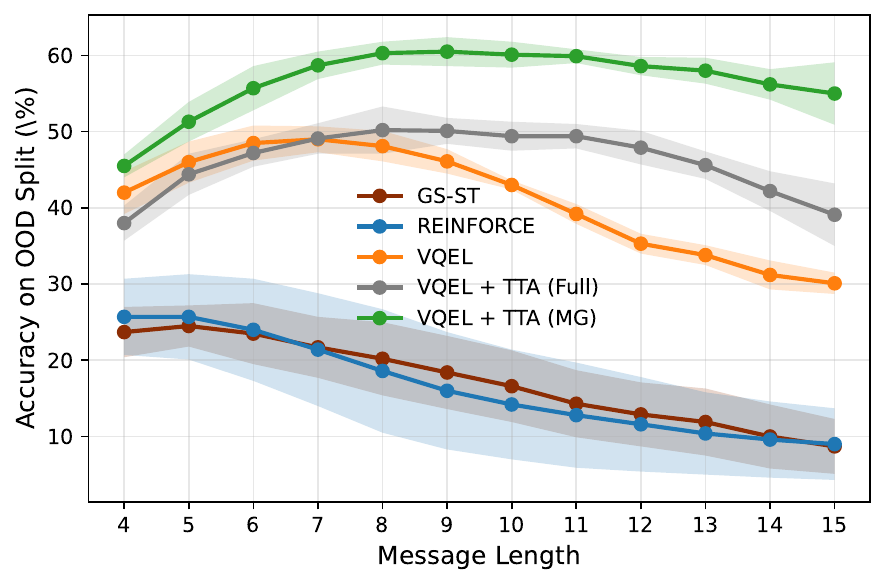}
    \end{minipage}
    \hfill
    \begin{minipage}{0.48\textwidth}
        \centering
        \includegraphics[width=\linewidth]{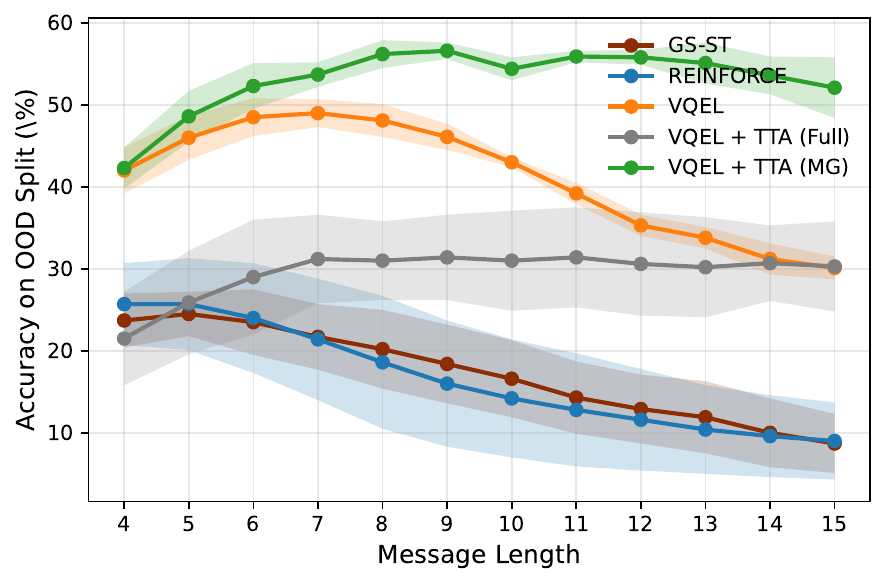}
    \end{minipage}
    \caption{\rev{Ablation study comparing our proposed TTA (adapting only the Message Generator) against Full Sender Adaptation on the MNIST dataset. Left: Batch-level; Right: Dataset-level.}}
    \label{fig:full-adapt-mnist}
\end{figure*}

\begin{figure*}[htbp]
    \centering
    \begin{minipage}{0.48\textwidth}
        \centering
        \includegraphics[width=\linewidth]{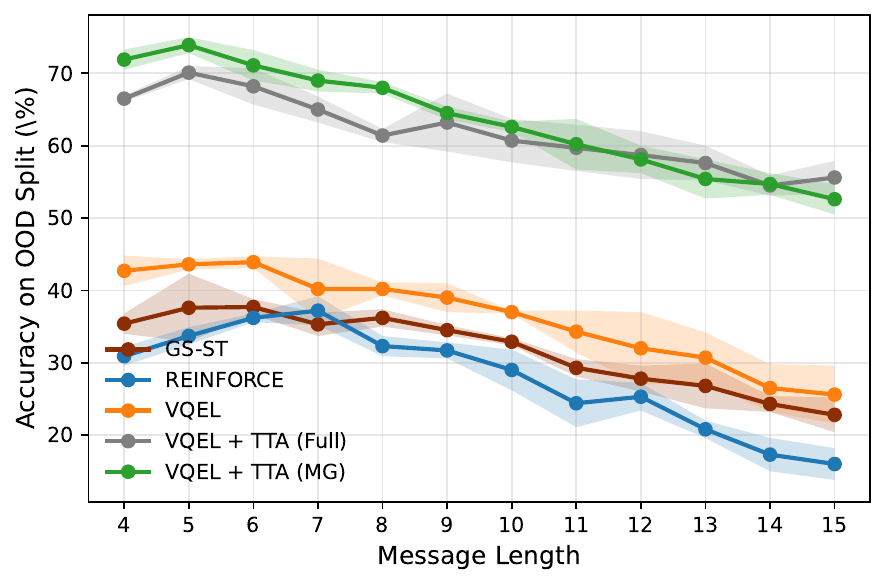}
    \end{minipage}
    \hfill
    \begin{minipage}{0.48\textwidth}
        \centering
        \includegraphics[width=\linewidth]{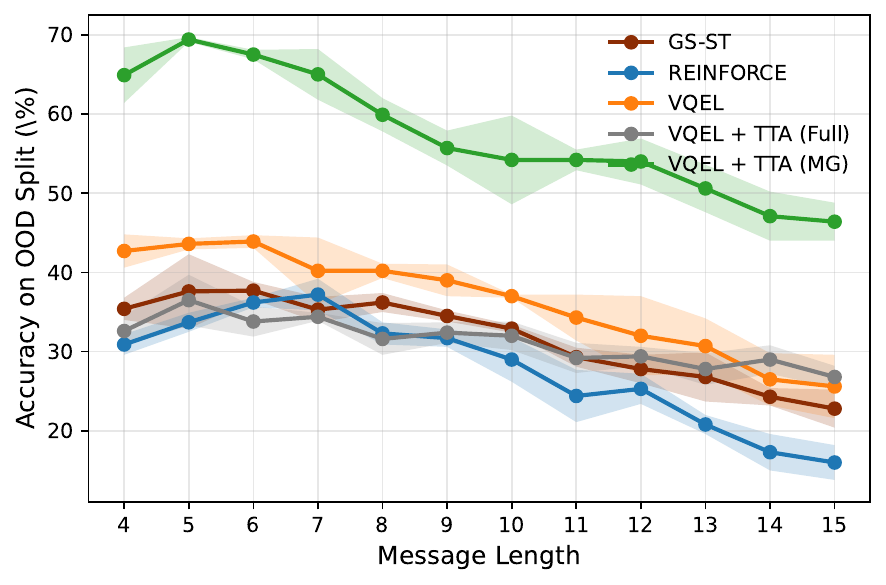}
    \end{minipage}
    \caption{\rev{Ablation study comparing our proposed TTA against Full Sender Adaptation on the ImageNet dataset. Left: Batch-level; Right: Dataset-level.}}
    \label{fig:full-adapt-imagenet}
\end{figure*}

\rev{Figures \ref{fig:full-adapt-mnist} and \ref{fig:full-adapt-imagenet} illustrate the performance comparison on the MNIST and ImageNet datasets, respectively. As observed, fine-tuning the entire sender either yields highly unstable results or severely degrades communication accuracy compared to adapting only the Message Generator. This empirical finding corroborates our theoretical justification in Section \ref{sec:limitations}: updating the perception modules at deployment without ground-truth receiver feedback causes the internal verifier to drift away from the categorization boundaries established during mutual play. By keeping the perception modules frozen, our method ensures the optimization target remains anchored, preventing representation collapse and maintaining alignment with the receiver.}

\subsection{\rev{Effect of the Random Length Trick}}
\label{app:random-length-trick}

\begin{figure*}
    \centering
    \begin{minipage}{0.48\textwidth}
        \centering
        \includegraphics[width=\linewidth]{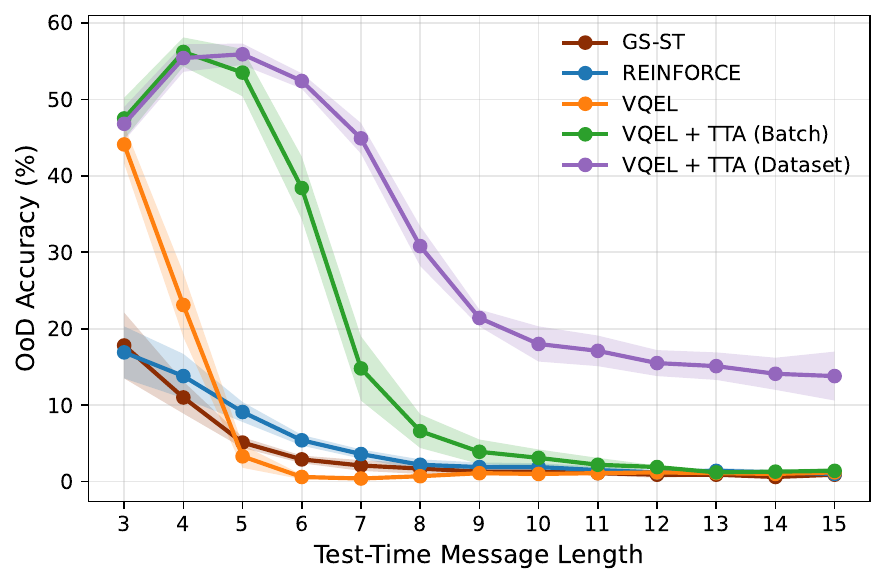}
    \end{minipage}
    \hfill
    \begin{minipage}{0.48\textwidth}
        \centering
        \includegraphics[width=\linewidth]{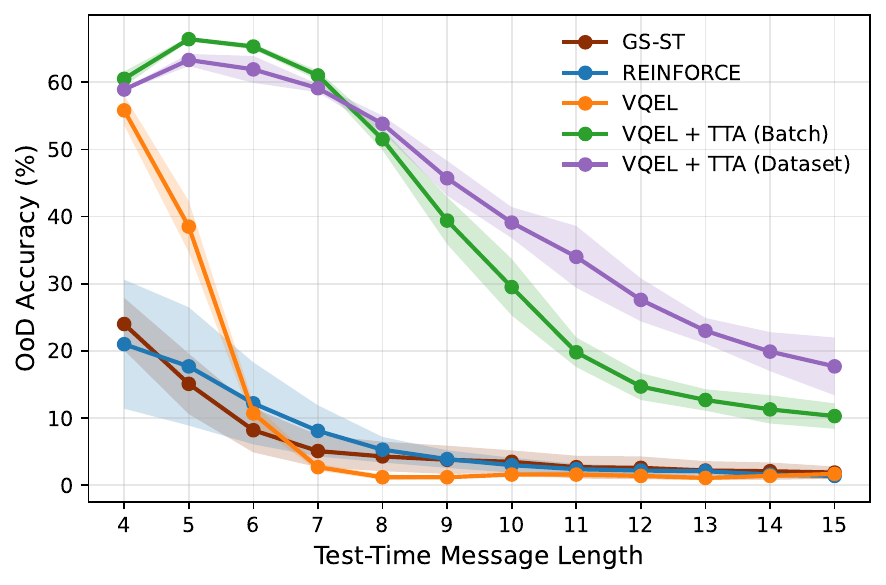}
    \end{minipage}
    \caption{\rev{OoD accuracy as a function of test message length on MNIST when RLT is removed. Left: models trained with fixed length $L=3$. Right: models trained with fixed length $L=4$. Accuracy degrades more rapidly for longer messages compared to Figure~\ref{fig:multi-object-complexity} (left), indicating reduced length generalization.}}
    \label{fig:mnist-without-rlt}
\end{figure*}

\begin{figure*}
    \centering
    \begin{minipage}{0.48\textwidth}
        \centering
        \includegraphics[width=\linewidth]{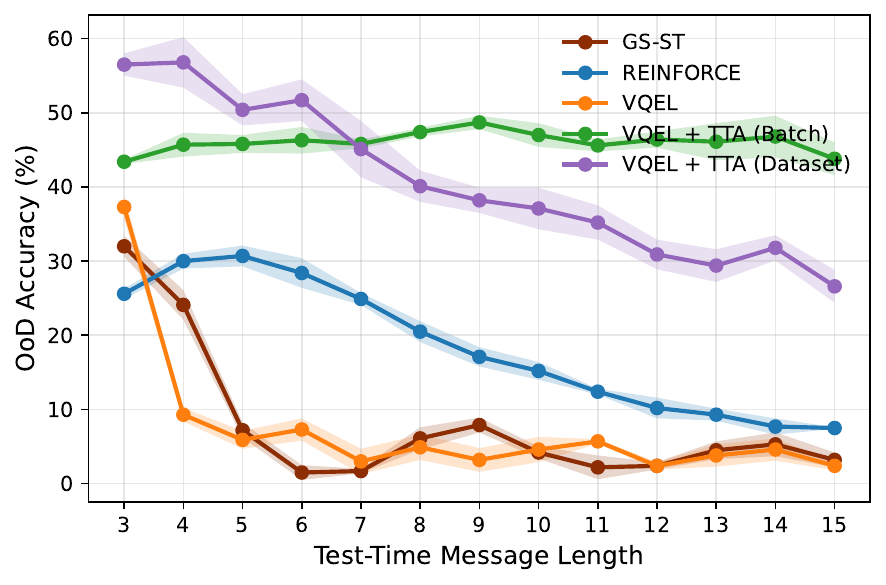}
    \end{minipage}
    \hfill
    \begin{minipage}{0.48\textwidth}
        \centering
        \includegraphics[width=\linewidth]{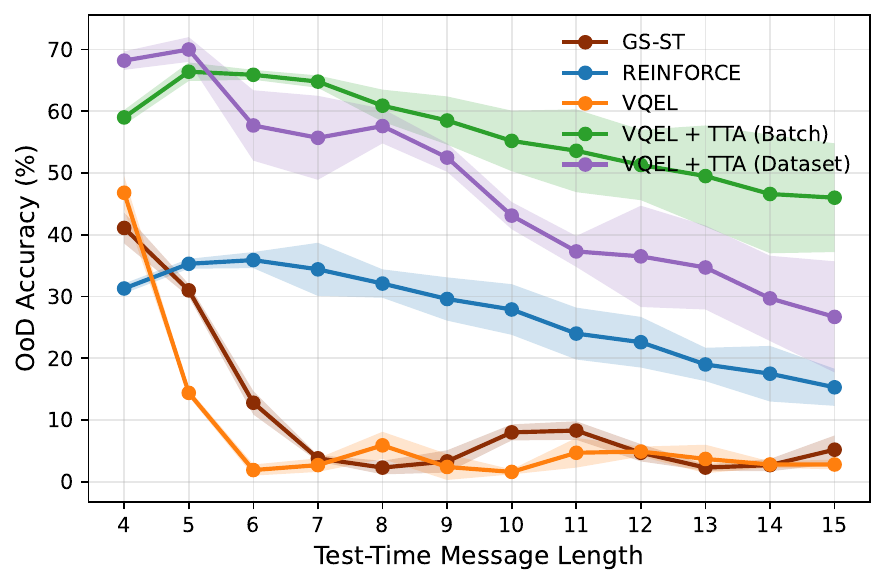}
    \end{minipage}
    \caption{\rev{OoD accuracy as a function of test message length on ImageNet when RLT is removed. Left: models trained with fixed length $L=3$. Right: models trained with fixed length $L=4$. Compared to Figure~\ref{fig:object-descriptive-complexity} (left), accuracy exhibits a steeper degradation with increasing message length, indicating weaker length generalization.}}
    \label{fig:imagenet-without-rlt}
\end{figure*}

\rev{
To improve robustness to variations in message length, all methods employ a \emph{Random Length Trick} (RLT) during training, where the message length for each training batch is randomly sampled from ${L-1, L}$. This introduces a limited degree of length variability during training while preserving the overall training procedure.

To assess the contribution of RLT, we trained ablated variants using a fixed message length of either $L$ or $L-1$ throughout training and evaluated them across varying message lengths at test time. Because RLT produces an average training message length of $(2L-1)/2$ (3.5 in our experiments), its performance should be interpreted relative to the fixed-length $L-1$ and $L$ baselines, between which its effective training budget lies.

Tables~\ref{tab:multi-object-complexity-without-rlt} and~\ref{tab:object-descriptive-complexity-without-rlt} report the resulting OoD accuracy degradation. The results suggest that the observed decline in OoD performance is not attributable to RLT; rather, it appears to be an inherent limitation of the evaluated methods.

\newrev{Figures~\ref{fig:mnist-without-rlt} and~\ref{fig:imagenet-without-rlt} illustrate OoD accuracy as a function of message length. The results demonstrate that removing RLT consistently reduces length generalization across all methods. Although models trained with a fixed message length achieve competitive performance near their training regime, their accuracy declines substantially more rapidly as the test message length increases. In contrast, models trained with RLT maintain higher accuracy on longer messages, indicating that RLT is an important supporting training-time component for robustness to length mismatch and extrapolation to unseen message lengths. The principal comparisons apply RLT uniformly to all methods, so they isolate the additional effect of TTA under this shared training protocol.}
}

\begin{table*}  
    \centering
    \setlength{\tabcolsep}{7pt}
    \renewcommand{\arraystretch}{1.1}

    \begin{tabular}{lcccc}
        \toprule
        \multirow{2}{*}{\textbf{Method}} 
        & \multicolumn{2}{c}{\textbf{L = 3}} 
        & \multicolumn{2}{c}{\textbf{L = 4}} \\
        \cmidrule(lr){2-3} \cmidrule(lr){4-5}
        & \textbf{1Digit (ID)} & \textbf{2Digit (OoD)}
        & \textbf{1Digit (ID)} & \textbf{2Digit (OoD)} \\
        \midrule
        GS-ST     & $69.1_{\pm 6.0}$ & $17.8_{\pm 4.3}$ & $83.3_{\pm 0.8}$ & $24.0_{\pm 3.9}$ \\
        REINFORCE & $75.9_{\pm 3.6}$ & $16.9_{\pm 3.4}$ & $85.8_{\pm 3.5}$ & $21.0_{\pm 9.6}$ \\
        VQEL      & $77.4_{\pm 0.7}$ & $44.1_{\pm 2.5}$ & $91.2_{\pm 0.6}$ & $55.8_{\pm 2.2}$ \\
        \bottomrule
    \end{tabular}
    \caption{\rev{Ablation of RLT on MNIST. Models are trained with a fixed message length ($L=3$ or $L=4$) rather than sampling from ${L-1,L}$. Results report ID accuracy on the training distribution and OoD accuracy. Consistent with the RLT setting, a substantial gap remains between ID and OoD performance.}}
    \label{tab:multi-object-complexity-without-rlt}
\end{table*}

\begin{table*}
    \centering
    \setlength{\tabcolsep}{7pt}
    \renewcommand{\arraystretch}{1.1}

    \begin{tabular}{lcccc}
        \toprule
        \multirow{2}{*}{\textbf{Method}} 
        & \multicolumn{2}{c}{\textbf{L = 3}} 
        & \multicolumn{2}{c}{\textbf{L = 4}} \\
        \cmidrule(lr){2-3} \cmidrule(lr){4-5}
        & \textbf{Standard (ID)} & \textbf{Complex (OoD)}
        & \textbf{Standard (ID)} & \textbf{Complex (OoD)} \\
        \midrule
        GS-ST     & $86.6_{\pm 0.5}$ & $32.0_{\pm 1.4}$ & $92.3_{\pm 0.5}$ & $41.1_{\pm 2.5}$ \\
        REINFORCE & $80.9_{\pm 0.5}$ & $25.6_{\pm 0.7}$ & $86.8_{\pm 1.3}$ & $31.3_{\pm 0.8}$ \\
        VQEL      & $83.9_{\pm 1.0}$ & $37.3_{\pm 1.6}$ & $91.9_{\pm 0.7}$ & $46.8_{\pm 2.7}$ \\
        \bottomrule
    \end{tabular}
    \caption{\rev{Ablation of RLT on ImageNet. Models are trained with a fixed message length ($L=3$ or $L=4$) rather than sampling from ${L-1,L}$. Results report ID accuracy on the Standard split and OoD accuracy on the Complex split. As in MNIST, a pronounced performance gap is observed between ID and OoD evaluations.}}
    \label{tab:object-descriptive-complexity-without-rlt}
\end{table*}

\section{Scaling at Test Time}
\label{sec:appendix-b}
As an alternative test-time strategy, we apply sampling-based scaling (TTS) on the OoD test splits. Specifically, for each test sample, we generate 50 candidate messages and select the one that minimizes the negative similarity between the object representation and the corresponding message representation:
\begin{equation}
\mathcal{L}_{\text{scaling}}(o)
=
 - \mathrm{sim}\!\left(
\mathbf{v}^{(S)}_m, \mathbf{v}^{(S)}_o
\right),
\label{eq:scaling-objective}
\end{equation}

\begin{figure*}
    \centering
    \begin{minipage}{0.48\textwidth}
        \centering
        \includegraphics[width=\linewidth]{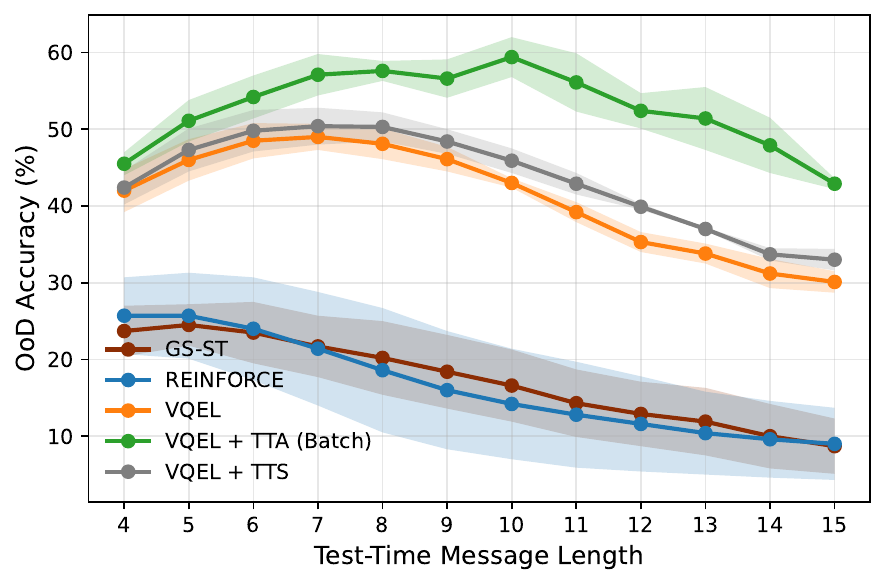}
    \end{minipage}
    \hfill
    \begin{minipage}{0.48\textwidth}
        \centering
        \includegraphics[width=\linewidth]{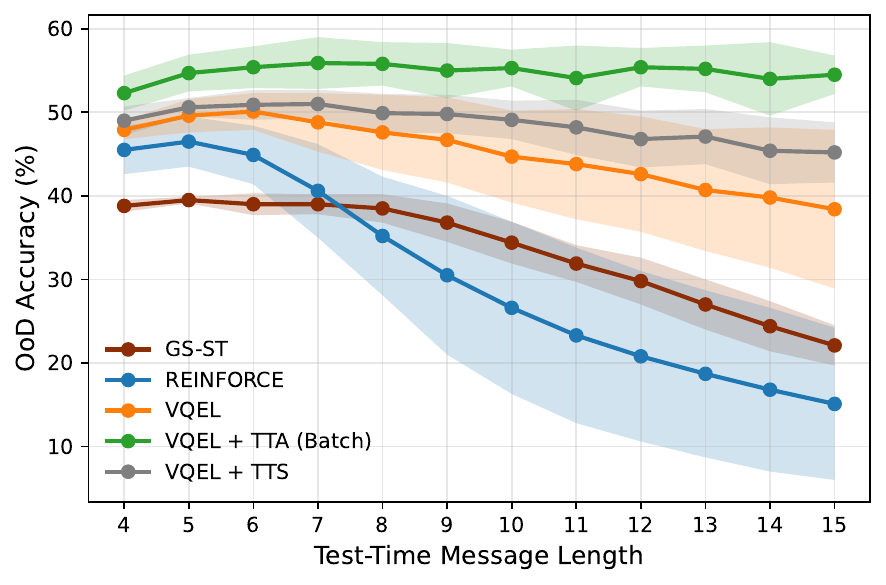}
    \end{minipage}
    \caption{\rev{TTS accuracy on the OoD split as a function of message length for datasets characterized by multi-object complexity. Left: MNIST; right: ShapeWorld.}}
  \label{fig:multi-object-complexity-scaling}
\end{figure*}

\begin{figure*}
    \centering
    \begin{minipage}{0.48\textwidth}
        \centering
        \includegraphics[width=\linewidth]{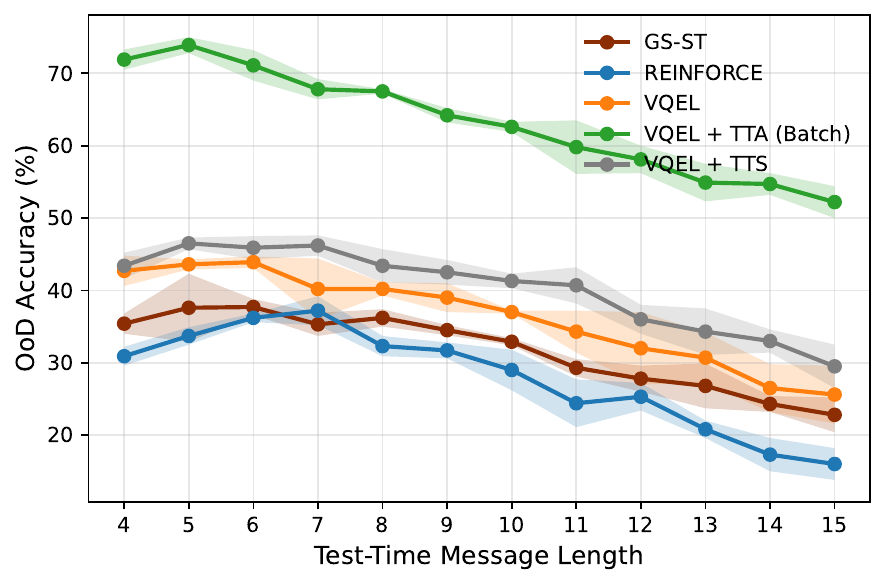}
    \end{minipage}
    \hfill
    \begin{minipage}{0.48\textwidth}
        \centering
        \includegraphics[width=\linewidth]{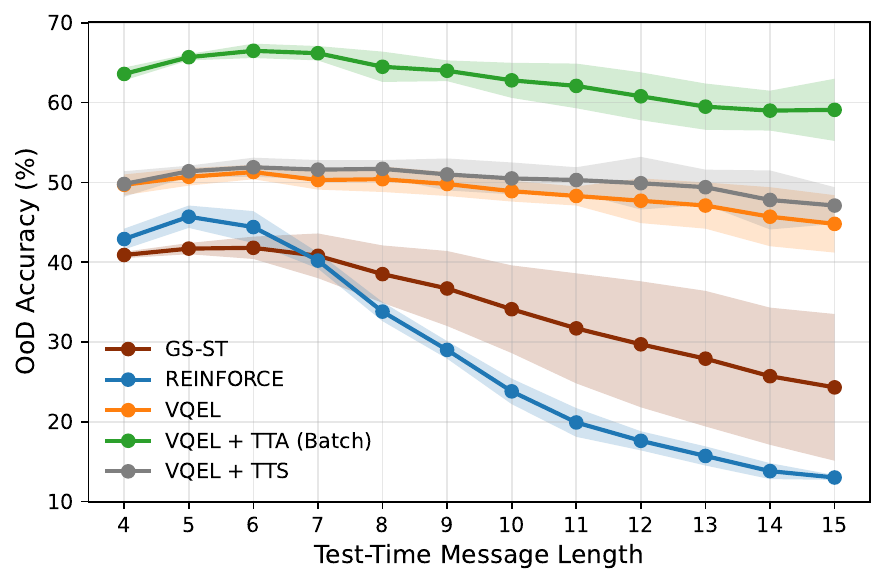}
    \end{minipage}
    \caption{\rev{TTS accuracy on the OoD split as a function of message length for datasets characterized by object-descriptive complexity. Left: ImageNet; right: COCO.}}
  \label{fig:object-descriptive-complexity-scaling}
\end{figure*}

We tune the sampling temperature over the set $\{10^{-2}, 10^{-1}\}$.
\rev{Figures \ref{fig:multi-object-complexity-scaling} and \ref{fig:object-descriptive-complexity-scaling} present the results across all four datasets. Consistent improvements are observed for all datasets and message lengths, with scaling typically increasing OoD test accuracy by approximately 2--6\%.}

\section{\rev{Qualitative Analysis of Messages Before and After TTA}}
\label{app:qualitative-analysis}

\begin{figure}[t]
    \centering
  \includegraphics[width=\linewidth]{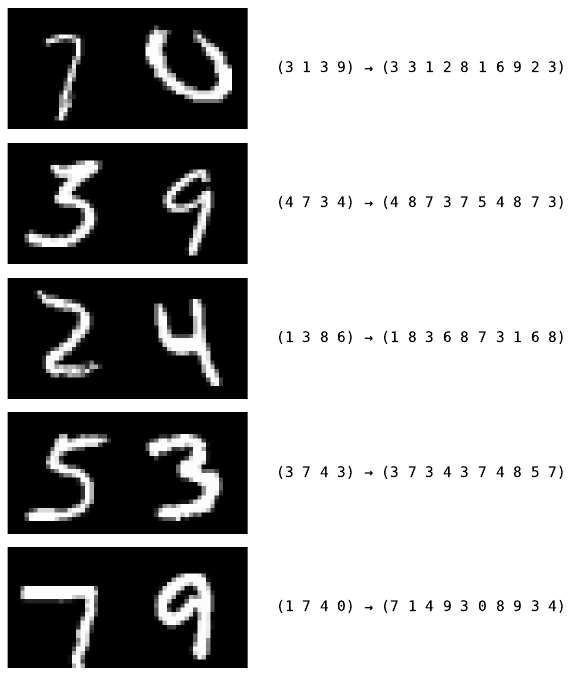}
        \caption{\rev{Qualitative examples of message adaptation on MNIST. The left column shows the input samples, and the right column illustrates the evolution of the learned messages before and after TTA (before $\rightarrow$ after).}}
  \label{fig:qualitative-examples-mnist}
\end{figure}

\begin{figure}[t]
    \centering
  \includegraphics[width=\linewidth]{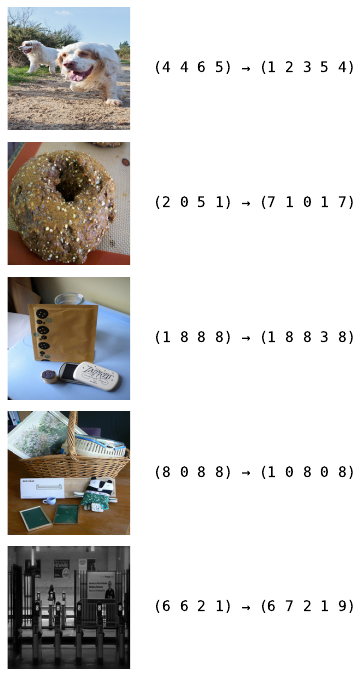}
    \caption{\rev{Qualitative examples of message adaptation on ImageNet. The left column shows the input samples, and the right column illustrates the evolution of the learned messages before and after TTA (before $\rightarrow$ after).}}
  \label{fig:qualitative-examples-imagenet}
\end{figure}

\rev{Figures \ref{fig:qualitative-examples-mnist} and \ref{fig:qualitative-examples-imagenet} show qualitative examples of the messages produced before and after TTA for five randomly sampled images from MNIST and ImageNet, respectively. The results reveal two distinct adaptation behaviors. For some samples, TTA preserves portions of the original message while extending it with additional tokens, indicating that the adapted protocol can incorporate new information to improve sample distinguishability. For other samples, the entire message representation changes after adaptation, suggesting that the model is free to reorganize its communication strategy rather than maintaining a fixed correspondence between tokens and concepts. This observation highlights that the emergent language learned by the model does not necessarily follow the structure of natural languages; instead, token semantics are task-dependent and may dynamically evolve during adaptation.}

\section{ImageNet OoD Splits}
\label{app:imagenet-ood-splits}

For our main experiments, we construct the ImageNet out-of-distribution (OoD) split by selecting distractor images from the same class as the target image. To study the effect of varying the number of distractors (Section \ref{sec:analysis-number-of-distractors}), we instead choose distractors whose embeddings are most similar to that of the target, due to the limited number of images available per class.

For this OoD configuration, we sample images from a subset of ImageNet-1k consisting exclusively of dog breeds—118 classes in total. Consequently, each candidate set contains only dog images, though each image corresponds to a distinct dog breed.
\section{\rev{Why CKA is Appropriate for Comparing Message Perception Representations}}
\label{app:cka_justification}
\rev{
The sender's internal Message Perception Module and the receiver's Message Perception Module are optimized independently and may therefore develop different coordinate systems within the same latent embedding space. Consequently, direct feature-wise comparisons can be misleading because two representations may encode the same information while expressing it through different combinations of latent features.

\newrev{To assess whether both modules encode similar representational structure, we employ CKA, a widely used measure of representational similarity. Rather than comparing individual feature dimensions, CKA compares the geometric structure of the representation spaces induced by the samples.}

\newrev{This property makes CKA particularly suitable for our setting. If two modules encode similar semantic structure but differ by an orthogonal transformation of the embedding space, such as a rotation, reflection, or permutation of feature dimensions, their representational geometry remains unchanged. CKA can therefore detect similar geometry even when there is no direct correspondence between individual latent dimensions. It is a representational diagnostic, not a direct test of identical decisions or OoD functional agreement.}
}

\section{\rev{Statistical Significance Assessment of CKA Scores}}
\label{app:cka_baseline}

\rev{
To determine whether the observed CKA similarity exceeds what would be expected by chance, we performed a permutation test to construct an empirical null distribution and compute a statistical significance level.

Let $\mathbf{X}\in\mathbb{R}^{N\times d_x}$ denote the representations produced by the sender's Message Perception Module and $\mathbf{Y}\in\mathbb{R}^{N\times d_y}$ denote the corresponding representations produced by the receiver's Message Perception Module, where each row corresponds to a message generated from a particular sample.

The observed similarity is computed as

$$
\mathrm{CKA}(\mathbf{X},\mathbf{Y}).
$$

To estimate a chance-level similarity distribution, we repeatedly destroy the correspondence between samples by randomly permuting the rows of $\mathbf{Y}$. Specifically, for each permutation $\pi$, we compute

$$
\mathrm{CKA}(\mathbf{X}, \pi(\mathbf{Y})).
$$

Randomly permuting the samples breaks the correspondence between sender and receiver representations while preserving the representations themselves. Therefore, any remaining similarity reflects chance-level alignment rather than meaningful representational agreement.

We perform 1000 independent permutations to obtain a null distribution of CKA scores. The reported random baseline corresponds to the mean and standard deviation of this null distribution. Statistical significance is assessed using the empirical permutation p-value

\begin{equation}
p = \frac{1 + \sum_{i=1}^{K}
\mathbb{I}\left[
\mathrm{CKA}_i^{\mathrm{perm}}
\ge
\mathrm{CKA}^{\mathrm{obs}}
\right]}
{K+1},
\end{equation}

where $K=1000$ is the number of permutations, $\mathrm{CKA}^{\mathrm{obs}}$ is the observed similarity, and $\mathrm{CKA}_i^{\mathrm{perm}}$ denotes the CKA score obtained from the $i$-th permutation. This procedure provides a non-parametric significance test requiring no distributional assumptions.
}
\section{\rev{Comparison of Oracle Adaptation Scopes}}
\label{app:oracles}
\begin{figure*}[htbp]
    \centering
    \begin{minipage}{0.48\textwidth}
        \centering
        \includegraphics[width=\linewidth]{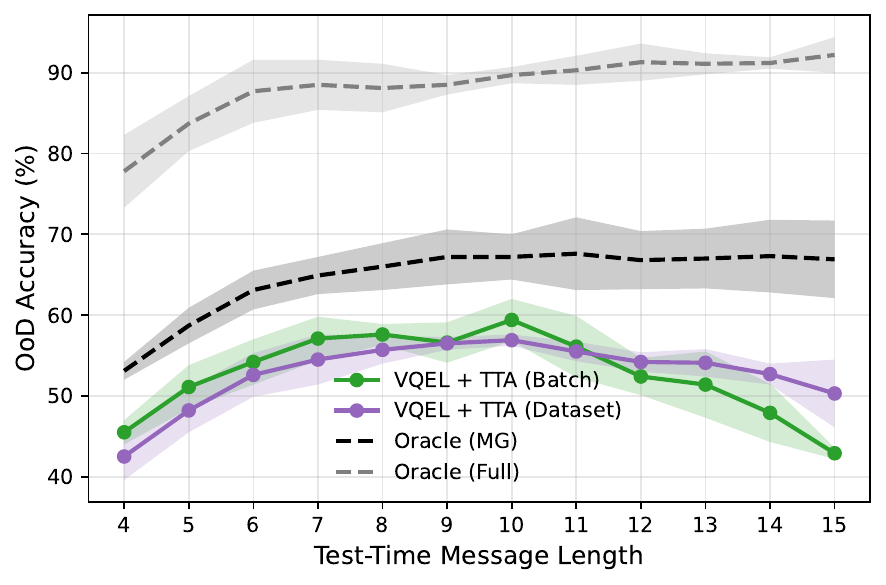}
    \end{minipage}
    \hfill
    \begin{minipage}{0.48\textwidth}
        \centering
        \includegraphics[width=\linewidth]{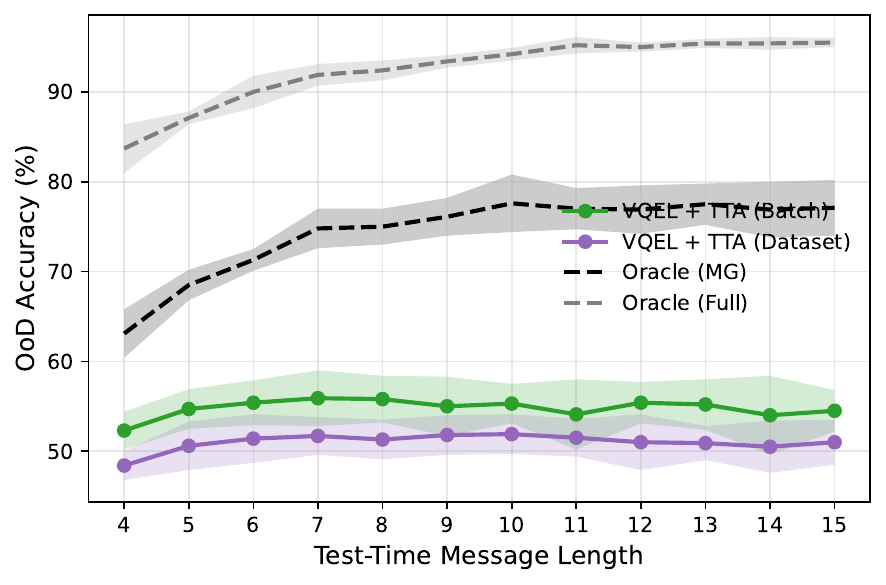}
    \end{minipage}
    \caption{\rev{Comparison between the compute-matched oracle used in the main paper and a stronger oracle that updates all sender and receiver components during OoD adaptation. The TTA methods are included for reference. Left: MNIST; Right: ShapeWorld.}}

    \label{fig:multi-object-complexity-oracles}
\end{figure*}

\begin{figure*}[htbp]
    \centering
    \begin{minipage}{0.48\textwidth}
        \centering
        \includegraphics[width=\linewidth]{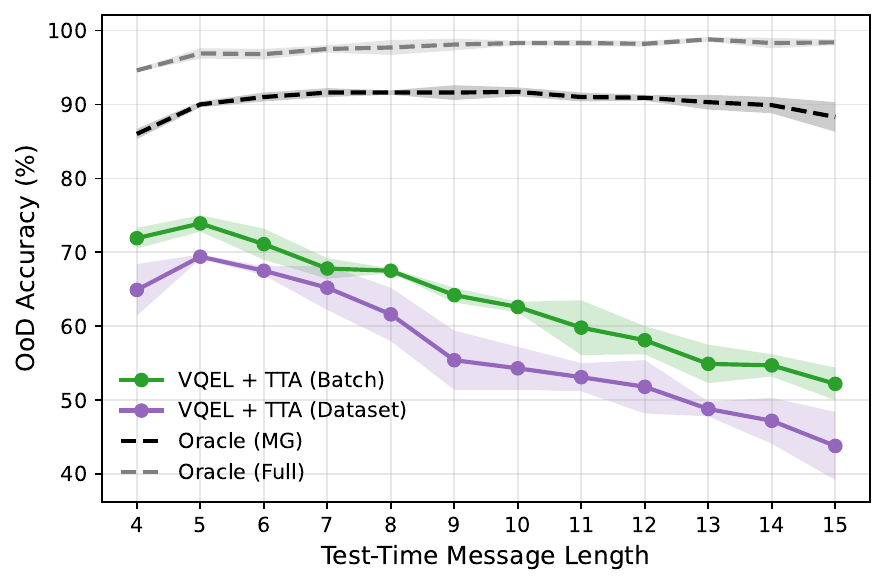}
    \end{minipage}
    \hfill
    \begin{minipage}{0.48\textwidth}
        \centering
        \includegraphics[width=\linewidth]{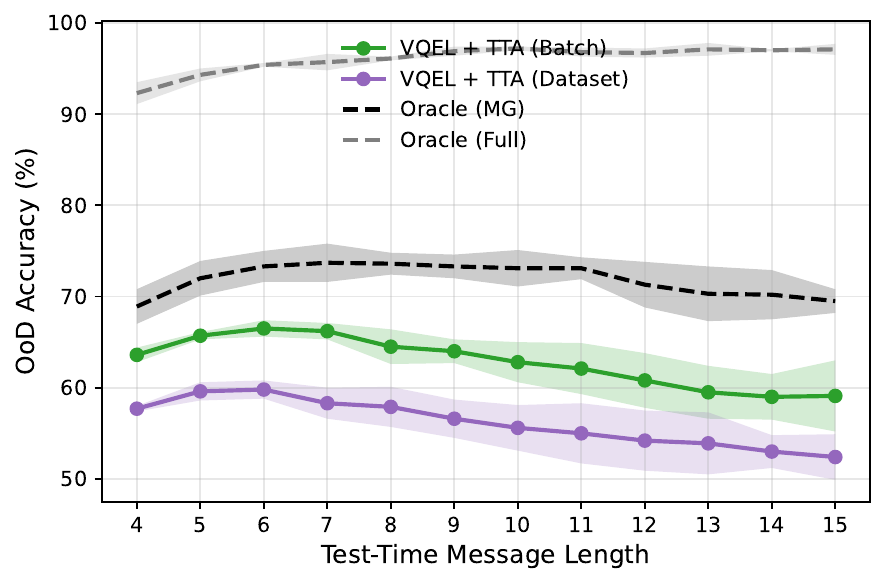}
    \end{minipage}
    \caption{\rev{Comparison between the compute-matched oracle used in the main paper and a stronger oracle that updates all sender and receiver components during OoD adaptation. The TTA methods are included for reference. Left: ImageNet; Right: COCO.}}

    \label{fig:object-descriptive-complexity-oracles}
\end{figure*}

\rev{
We additionally evaluate a stronger oracle that updates all sender and receiver components during OoD adaptation using the receiver loss. The results are shown in Figures~\ref{fig:multi-object-complexity-oracles} and~\ref{fig:object-descriptive-complexity-oracles}. Unlike the oracle used in the main paper, which is restricted to updating only the Message Generator during adaptation, this variant updates all sender and receiver components and therefore represents a less restrictive upper bound.
}

\FloatBarrier

\bibliographystyle{IEEEtran}
\bibliography{custom}

\newpage


     
\begin{IEEEbiography}[{\includegraphics[width=1in,height=1.25in,clip,keepaspectratio]{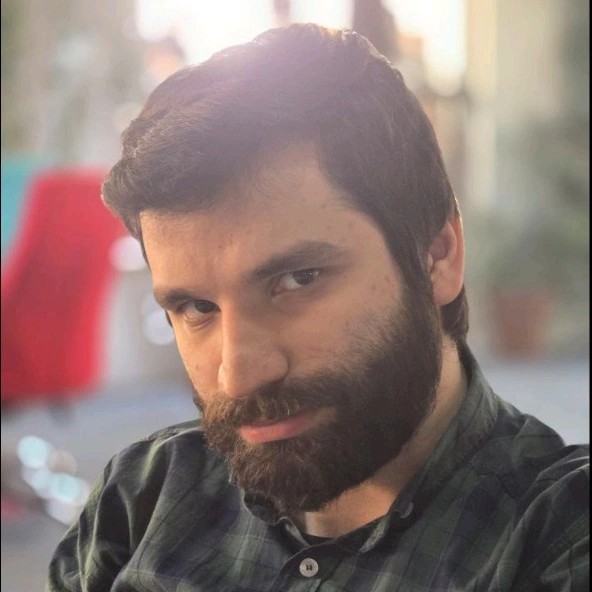}}]{Mohammad Mahdi Samiei}
    was born in Tehran, Iran, in 1997. He received the B.Sc. degree in computer engineering from Amirkabir University of Technology, Tehran, in 2018, and the M.Sc. degree in artificial intelligence from Sharif University of Technology, Tehran, in 2020. He is currently pursuing the Ph.D. degree in artificial intelligence. His doctoral research focuses on compositional inductive biases for out-of-distribution (\newrev{OoD}) generalization. Accordingly, his primary research interests include the study of emergent languages and \rev{Complexity-OoD} scenarios.   
    \end{IEEEbiography}

\begin{IEEEbiography}[{\includegraphics[width=1in,height=1.25in,clip,keepaspectratio]{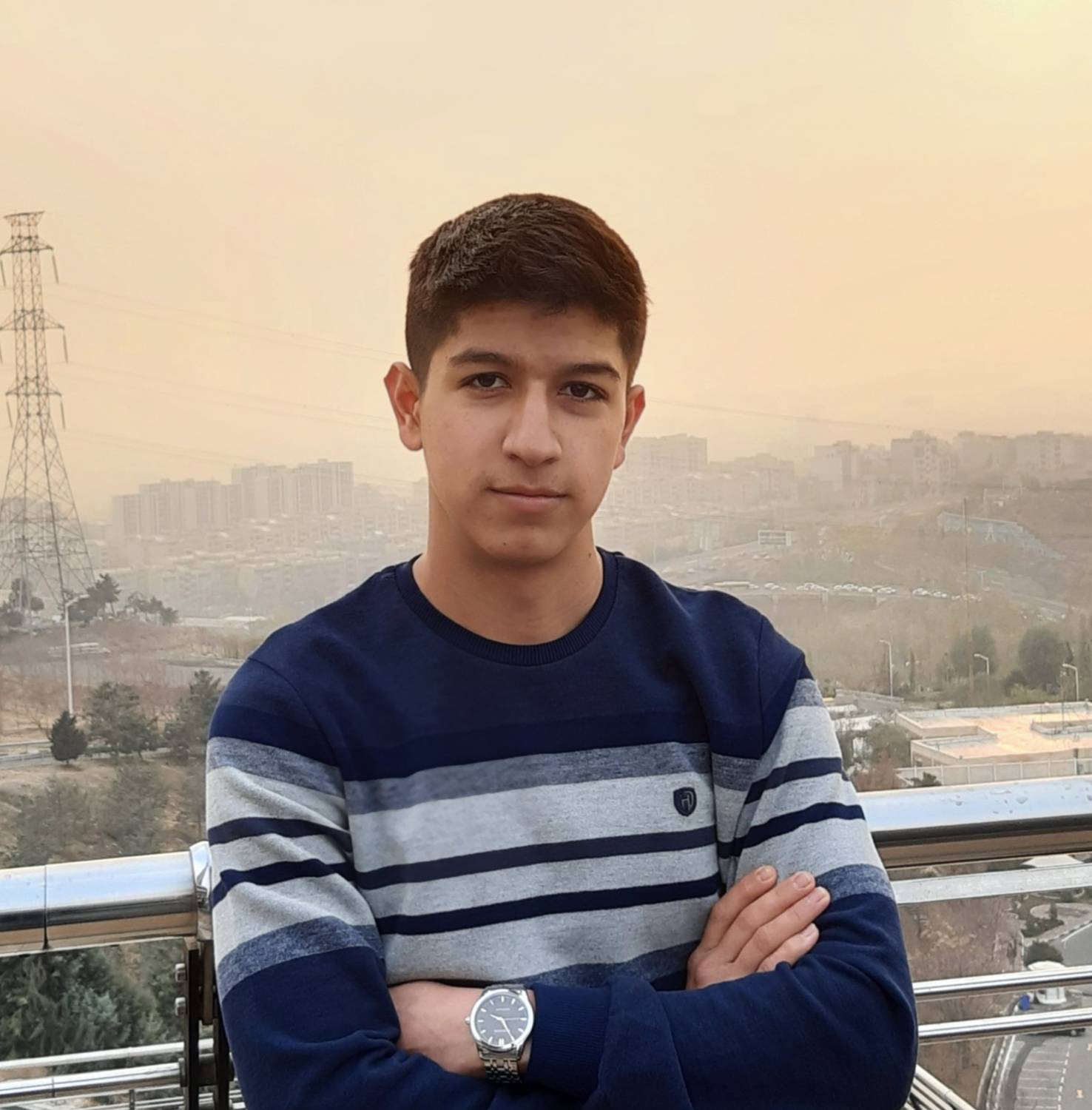}}]{Mehdi Jamalkhah}
\rev{
is currently pursuing the B.Sc. degree in computer engineering at the University of Tehran, Tehran, Iran. 
His research interests include large language models, with a focus on creativity and reasoning capabilities, as well as deep learning, particularly vision-language representation learning.
}
\end{IEEEbiography}

\begin{IEEEbiography}[{\includegraphics[width=1in,height=1.25in,clip,keepaspectratio]{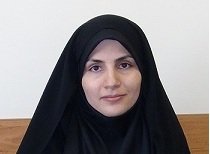}}]{Mahdieh Soleymani Baghshah}
degrees from the Department of Computer Engineering, Sharif University of
Technology, Iran, in 2003, 2005, and 2010, respectively. Her M.S. and Ph.D.
theses were in the field of machine learning. She joined the Sharif University
of Technology as an Assistant Professor of computer engineering, where
she has founded the Machine Learning Laboratory (MLL), in 2012. She is
currently an Associate Professor with the Sharif University of Technology
and the Director of MLL. Her research interests include machine learning
and deep learning.
\end{IEEEbiography}

\EOD

\end{document}